\documentclass[a4paper,fleqn]{cas-dc}

\usepackage[numbers]{natbib}
\usepackage{subfigure}
\usepackage[center]{caption}
\usepackage[switch]{lineno}

\def\tsc#1{\csdef{#1}{\textsc{\lowercase{#1}}\xspace}}
\tsc{WGM}
\tsc{QE}
\tsc{EP}
\tsc{PMS}
\tsc{BEC}
\tsc{DE}
\begin{document}
\let\WriteBookmarks\relax
\def\floatpagepagefraction{1}
\def\textpagefraction{.001}

% Short title
\shorttitle{Intelligent Agricultural Recommendation System}

% Short author
\shortauthors{Zubair et~al.}

% Main title of the paper
%\title [mode = title]{Agricultural Recommendation System based on Multivariate Weather Forecasting Model}   
% Main title of the paper
\title [mode = title]{Agricultural Recommendation System based on Deep Learning: A Multivariate Weather Forecasting Approach} 
% eg: \tnotemark[1]
% \tnotemark[1,2]

% Title footnote 1.
% eg: \tnotetext[1]{Title footnote text}
% \tnotetext[<tnote number>]{<tnote text>} 
%\tnotetext[1]{This document is the results of the research project funded by the National Science Foundation.}

%\tnotetext[2]{The second title footnote which is a longer text matter to fill through the whole text width and overflow into another line in the footnotes area of the first page.}

% First author
%
% Options: Use if required
% eg: \author[1,3]{Author Name}[type=editor,
%       style=chinese,
%       auid=000,
%       bioid=1,
%       prefix=Sir,
%       orcid=0000-0000-0000-0000,
%       facebook=<facebook id>,
%       twitter=<twitter id>,
%       linkedin=<linkedin id>,
%       gplus=<gplus id>]
\author[1]{Md Zubair}[orcid=0000-0001-7384-2805]

% Corresponding author indication
\cormark[1]

% Footnote of the first author
% \fnmark[1]

% Email id of the first author
\ead{zubairhossain773@gmail.com}

% % URL of the first author
% \ead[url]{www.cvr.cc, cvr@sayahna.org}

%  Credit authorship
\credit{Conceptualization of this study, Methodology, Software}

% Address/affiliation
\affiliation[1]{organization={Department of Computer Science and Engineering, Chittagong University of Engineering \& Technology},
    addressline={Chattogram-4349}, 
    % city={Amsterdam},
    % citysep={}, % Uncomment if no comma needed between city and postcode
    % postcode={1043 NX}, 
    % state={},
    country={Bangladesh}}

% Second author
\author[2]{Md. Shahidul Salim}
% Address/affiliation
\affiliation[2]{organization={Department of Computer Science and Engineering, Khulna University of Engineering \& Technology},
    % addressline={}, 
    city={Khulna-9203},
    % citysep={}, % Uncomment if no comma needed between city and postcode
    % postcode={695014}, 
    % state={Trivandrum},
    country={Bangladesh}}

% Third author
\author[3]{Mehrab Mustafy Rahman}
% \fnmark[2]
% \ead{cvr3@sayahna.org}
% \ead[URL]{www.sayahna.org}
\affiliation[3]{organization={Department of Computer Science, University of Illinois at Chicago},
    addressline={Chicago, IL 60607}, 
    % city={Malayinkil},
    % citysep={}, % Uncomment if no comma needed between city and postcode
    % postcode={695571}, 
    % state={Trivandrum},
    country={USA}}

\credit{Data curation, Writing - Original draft preparation}

% Fourth author
\author[1]{Mohammad Jahid Ibna Basher}
% \cormark[2]
% \fnmark[1,3]
% \ead{rishi@stmdocs.in}
% \ead[URL]{www.stmdocs.in}

\author[4]{Shahin Imran}
\affiliation[4]{organization={Department of Agronomy, Khulna Agricultural University},
    addressline={Khulna-9202}, 
    % city={Malayinkil},
    % citysep={}, % Uncomment if no comma needed between city and postcode
    % postcode={695571}, 
    % state={Trivandrum},
    country={Bangladesh}}

\author[5]{Iqbal H. Sarker}[orcid=0000-0003-1740-5517]
\cormark[1]
\ead{m.sarker@ecu.edu.au}
\affiliation[5]{organization={Centre for Securing Digital Futures, School of Science, Edith Cowan University},
    addressline={Perth, WA-6027}, 
    % city={Malayinkil},
    % citysep={}, % Uncomment if no comma needed between city and postcode
    % postcode={695571}, 
    % state={Trivandrum},
    country={Australia}}

% Corresponding author text
\cortext[cor1]{Corresponding author}
% \cortext[cor2]{Principal corresponding author}

% Footnote text
%\fntext[fn1]{This is the first author footnote. but is common to thirdauthor as well.}
%\fntext[fn2]{Another author footnote, this is a very long footnote and it should be a really long footnote. But this footnote is not yet sufficiently long enough to make two lines of footnote text.}

% For a title note without a number/mark
%\nonumnote{This note has no numbers. In this work, we demonstrate $a_b$ the formation Y\_1 of a new type of polariton on the interface between a cuprous oxide slab and a polystyrene micro-sphere placedon the slab.}

% Here goes the abstract
\begin{abstract}
Agriculture plays a fundamental role in driving economic growth and ensuring food security for populations around the world. Although labor-intensive agriculture has led to steady increases in food grain production in many developing countries, it is frequently challenged by adverse weather conditions, including heavy rainfall, low temperatures, and drought. These factors substantially hinder food production, posing significant risks to global food security. In order to have a profitable, sustainable, and farmer-friendly agricultural practice, this paper proposes a context-based crop recommendation system powered by a weather forecast model. For implementation purposes, we have considered the whole territory of Bangladesh.  \textit{With extensive evaluation, the multivariate Stacked Bi-LSTM (three Bi-LSTM layers with a time Distributed layer) Network is employed as the weather forecasting model.} The proposed weather model can forecast Rainfall, Temperature, Humidity, and Sunshine for any given location in Bangladesh with an average $R^2$ value of 0.9824, and the model outperforms other state-of-the-art LSTM models. These predictions guide our system in generating viable farming decisions. Additionally, our full-fledged system is capable of alerting the farmers about extreme weather conditions so that preventive measures can be undertaken to protect the crops. Finally, the system is also adept at making knowledge-based crop suggestions for flood and drought-prone regions.

\end{abstract}

% Use if graphical abstract is present
% \begin{graphicalabstract}
% \includegraphics{figs/grabs.pdf}
% \end{graphicalabstract}

% Keywords
% Each keyword is seperated by \sep

\maketitle

\section{Introduction}
% Reviewer 1 (Jahid)
% Update- add novelty(first introduce the model for bangladesh), result (abstract)
% Update - keywords need to be updated 
% Update - add limitation (more features, sampling rate (jahid), transformer based model for future work)
% update _ add some implication to the intro (https://link.springer.com/article/10.1007/s00521-017-2948-1, https://www.tandfonline.com/doi/full/10.1080/16168658.2021.1886813)
% Reviewer 3 (Zubair)
% Update _ clarification for the crop suggestion 
% Update - add details in the table captions of figure -11, fig -12 and so on.

% Reviewer - 4 
% Update - add details on the caption of the figure 13 (Zubair)
% Update - add figure against table 8 for recommendation. (Zubair)
% Update - add climate change issue in the intro and connect with SDG goals. 
%  update - APA and vancuver referencing should be removed (Zubair)

% update - divide the 5.1.2 into paragraphs (Zubair)
% Update - referencing should be revised (Zubair)

% Reviewer - 5
%  Update - size of the dataset should be mentioned (Zubair)
% update - training time should be added (Zubair)

Advancement in agriculture is essential for human survival to ensure global food security, economic growth, and sustainable development. Human civilization relies on the development of agriculture not only for its food supply but also for economic growth. A significant portion of global GDP comes from the agricultural sector. For developing and least developing countries, agriculture contributes more than 25\% of the GDP \cite{FAO, WorldBank}. In recent years, farmers have faced challenges in food production due to extreme weather conditions driven by climate change and limited resources (water, arable land, manpower, etc.) \cite{FAO1}. To tackle the situation, farmers need to stay informed about future weather conditions and their changing patterns so that they make appropriate agricultural decisions to address challenges and boost productivity in the agrifood sector.

The limited availability of advanced techniques and information for farmers and the labor force to manage the increasing impacts of climate change is a considerable issue. A reliable weather forecast according to the farmers' needs is crucial for making optimal farming decisions. However, in developing countries, the absence of such tools forces farmers to rely on traditional techniques, which often results in low yields of crops. Complications such as imbalanced use of fertilizers, inefficient use of water (irrigation), pests and diseases, and increased dependence on the mercy of nature are the primary reasons behind a low yield. These hindrances can easily be dealt with the explicit forecasting of weather and taking the necessary steps to deal with it.

There are some intelligent models to predict crop production and forecast weather \cite{cite7}. Moreover, Climate Smart Agriculture (CSA) techniques are also in existence to solve the problems. Use of submergence, resistant, high-yielding varieties, and solar-powered irrigation \cite{cite8} are some of the modern techniques. Other processes such as the Sorjan method, rainwater harvesting, flood tolerant crop varieties \cite{cite9}, efficient resource management, genetically modified crop culture, and developing forecasting system \cite{cite10} can also be utilized for combating the effect of climate change in agriculture. Additionally, alternate wetting and drying \cite{cite11} are some of the most innovative and effective approaches suggested. However, the real-world application of these technologies and techniques is very much limited and followed in only a few of the farms and arable land in developing countries. In most cases, farmers aren't aware of these technologies. So, we have proposed a reliable multivariate weather forecast model and tried to suggest modern techniques based on the forecast output. For implementation purposes, we have considered Bangladesh, one of the developing countries of South Asia.

% Accurate weather forecasting presents significant challenges due to its nature (time series data), where predictions rely heavily on both past and future data. Existing statistical models for weather prediction often fall short in capturing the intricate patterns of weather behavior.

In this research article, we work to address the above mentioned issues by creating an intelligent recommendation system based on a multivariate weather forecast model. Accurate weather forecasting is a challenging task due to the nature of its nature, where predictions rely on the past and future of time series data. Existing statistical models for weather prediction are not sufficient to capture the intricate patterns of the weather. Similarly, conventional supervised regression machine learning models aren't also susceptible to predicting weather efficiently. But Bi-directional long short-term memory can address these issues. Therefore, we have used the model and customized it for better performance according to our specific requirements. We faced many challenges in the data collection and preprocessing stages. The collected data was unstructured, containing a lot of missing values. However, we overcame the challenges. The purpose of this model will be to forecast the weather parameters: Rainfall, Humidity, Temperature, and Sunshine. Finally, the weather forecasting is done based on time series \cite{cite13,cite14} data using a bi-directional long short-term memory (Bi-LSTM) model \cite{cite12}. The main contributions of the research work are:
\begin{itemize}
    \item Collecting, structuring, and analyzing real-world weather data from the inception of every weather station of Bangladesh containing temperature, rainfall, humidity, sunshine, and wind direction features.
    \item Developing an efficient weather forecasting model for the farmers using a stacked Bi-LSTM model. For this, we modify the basic Bi-LSTM architecture for higher eﬀiciency.
    \item Providing context-based agricultural recommendations derived from the model's forecast and other parameters. 
\end{itemize}
The paper is organized in the following order. In Sect. \ref{related_work}, we discuss some relevant works that comply with the proposed architecture. In Sect. \ref{data_collection}, we describe the overall data collection and preprocessing process. The working principle of our proposed model and the evaluation process are discussed in Sect. \ref{proposed_methodology} and \ref{evaluation} respectively. We highlight some key observations of our study in Sect. \ref{discussion} and finally, Sect. \ref{conclusion} concludes this paper.

\section{Related Work}
\label{related_work}
Agriculture is one of the important sectors of a country which fulfills our very basic needs. Human civilization emerged based on agriculture \cite{gupta2004origin}. Many researchers and scientists are working tirelessly to integrate agriculture with modern techniques and technologies. We will discuss some of the works that align with our work.

Pradeep et al. propose a novel lightweight data-driven weather forecasting model by exploring temporal modeling approaches of long short-term memory (LSTM) and temporal convolutional networks (TCN) and compared its performance with existing classical machine learning approaches, statistical forecasting approaches, and a dynamic ensemble method, as well as the well-established weather research and forecasting (WRF) NWP model. The experiment reveals that the proposed LSTM with TCN model produces more accurate results than the well-known and complex WRF model \cite{hewage2021deep}.
In article \cite{karevan2020transductive}, the authors created a weather forecasting model using LSTM. They used transductive LSTM (T-LSTM) to predict time series using local information. Experiments are done twice a year to test the technique in different weather conditions. The Weather Underground website measured minimum and maximum temperature, dew point, and wind speed in Brussels, Antwerp, Liege, Amsterdam, and Eindhoven. Another simple model developed by Zubair et al. suggests the best crops based on user location in the context of Bangladesh. In case of suggestions, they consider temperature, rainfall, and other soil properties for predicting crop yield. Statistical method Seasonal Auto-Regressive Integrated Moving Average (SARIMA) was used for temperature and rainfall forecast. The model can only forecast a single value at a time and is a static model \cite{cite7}. Another group of researchers \cite{scher2021ensemble} have tested four distinct approaches to converting a deterministic neural network weather forecasting system into an ensemble forecasting system. They made use of the atmospheric data from ERA5, the most recent reanalysis output from ECMWF. It was found that a deterministic neural network forecast is less skillful than the ensemble mean of all methods. Continuing the same theme, Espeholt et al.\cite{espeholt2022deep} proposed neural network-based models - MetNet-2 and MetNet-2 Hybrid (combination of MetNet2 and HRRR models) that outperform state-of-the-art physics-based models both in prediction and computation. The models are capable of predicting precipitation at a high resolution up to 12 hours ahead of time. 

A Deep Q Learning irrigation decision-making technique was proposed by Chen et al.\cite{chen2021reinforcement}. Compared to regular irrigation decision-making, this technique has decreased water saving by 23mm and reduced drainage by 21mm. The experiment was verified on single-season rice and double-season rice in the Nanching region in Southern China. The mean reward for the early, middle, and late varieties of 2019 is 7.57, 8.70, and 9.56, respectively. The value of the reward function showed a specific correlation with rainfall. To predict the weather of a city in Austin, Texas, USA, Datta et al.\cite{datta2020complete} used supervised Machine Learning Algorithms - Artificial Neural Network and Gradient Boosting Algorithm. They have used average temperature, average dew point, average sea level pressure, average humidity, etc. The data contained 7 classes: rainy, not rainy, snowy, foggy, thunderstorm, rainy with the thunderstorm, fog. The accuracy of the Artificial Neural Network and Gradient Boosting models are 78.54\% and 78.92\%, respectively. Different researchers have also proposed some other systems using Numerical Weather Prediction (NWP) techniques. Haupt et al.\cite{haupt2018machine} discuss the DIcast System, an automated engine that uses NWP with
historical observations from that location. For short ranges, a system has been
developed using the K-means algorithm to identify cloud regimes; after that, an Artificial Neural Network (ANN) was used on each regime, which out-predicts any model using the regular ANN without clustering the cloud regimes. The National Center for Atmospheric Research performed the whole study. \textit{None of the above studies consider the important parameters (Rainfall, Temperature, Humidity, Sunshine Hours) and a sustainable forecasting model for the entire country. We will try to address the issue and solve the research gap.} 

Toni et al. review article talks about how farmers use weather and climate data to make decisions about their crops and livestock. They can use this data to decide when to plant and harvest, which crops to grow, and how to take care of their animals. However, weather forecasts can be difficult to trust because climate models are not perfect and can be different from each other. Other factors, such as money and land, also affect what farmers do. Overall, this review talks about how weather and climate information can be helpful for farmers, but they need to consider other things too \cite{klemm2017development}.
Amber et al. review 30 years of US, Australian, and Canadian studies on agricultural decision-makers' usage and perceptions of meteorological and climate information and DSTs. They reviewed many works to find study methods, climate data, agricultural changes, and farmer opinions and found that boosting prediction ”skill” or dependability would be useful, but convincing farmers of meteorological instrument precision and dependability is crucial\cite{mase2014unrealized}.

Many researchers are working on predicting the weather for agricultural purposes. A study shows the effect of long and short-term weather forecasts on agriculture in Europe. Different types of decisions, like irrigation, crop cultivation, etc., had a direct impact on agricultural decision-making. Additionally, there are concerns about the eﬀiciency and accuracy of long-term weather forecasting \cite{calanca2011application}.
S. Paparrizos et al. published an exciting and important article showing the necessity of weather forecasting only for peri-urban farmers of Bangladesh. They also mentioned how weather forecasting could help the farmers in case of profitable decision-making all year round \cite{paparrizos2020verification}. Unlike these works, we effectively build an agriculture recommendation model based on multivariate weather forecasting models by considering the above-mentioned importance and real-world applicability.

\section{Data Collection and Prepossessing}
Real-world weather data collection is the most crucial part of developing our proposed model. After collecting the data, we noticed that the data was not fully structured and contained missing values. So, we carefully analyzed the data to get in-depth insights into it and applied different techniques for preprocessing. In this section, we will discuss all the steps for data collection, analysis, and preprocessing in detail. 
\label{data_collection}
\subsection{Dataset collection}
Data is the fuel of a data-driven model. Our primary target is to create a weather forecasting model and generate agricultural decisions. We consider the entire area of Bangladesh for the experimental purpose of the research. Therefore, we have collected all the previous data from the weather stations (35 weather stations) in Bangladesh \cite{Weatherstation}. The weather data is not publicly available. We have purchased the data from the meteorological department of Bangladesh. The authority has handed over the data for non-commercial educational use only. They have provided data on Rainfall, Temperature, Humidity, Sunshine (Daily hours of sunshine), Wind Speed, and Wind Direction from the starting year of the stations to 2022. Table \ref{tab:table1} represents the duration of data collection, total number of instances and data type of the dataset.
\begin{table}[]
\begin{center}
\scalebox{0.8}{
\begin{tabular}{|l|c|c|c|c|}
\hline
\textbf{Station Location} & \begin{tabular}{@{}c@{}}\textbf{Starting} \\ \textbf{Year}
\end{tabular} & \begin{tabular}{@{}c@{}}  \textbf{Ending} \\ \textbf{Year}
\end{tabular} & \begin{tabular}{@{}c@{}} \textbf{Sample} \\ \textbf{Size} \end{tabular} & \begin{tabular}{@{}c@{}} \textbf{Data} \\ \textbf{Type} \end{tabular} \\ \hline
\begin{tabular}[c]{@{}l@{}}Barisal, Bogra, Chittagong, \\ Comilla, Cox'sBazar, Dhaka, \\ Dinajpur, Faridpur, Jessore, \\ Khulna, M.court, Mymensingh, \\ Rangamati, Rangpur, Satkhira, \\ Srimangal, Sylhet \end{tabular} & 1960 & \multirow{12}{*}{\rotatebox[origin=c]{90}{2022}} & \multirow{12}{*}{\rotatebox[origin=c]{90}{489268}} & \multirow{12}{*}{\rotatebox[origin=c]{90}{Time Series (Numerical, Categorical)}} \\ \cline{1-2}
Ishurdi & 1961 & & & \\ \cline{1-2}
Chandpur, Rajshahi & 1964 & & & \\ \cline{1-2}
Bhola, Hatiya & 1966 & & & \\ \cline{1-2}
Sandwip & 1967 & & & \\ \cline{1-2}
Patuakhali & 1973 & & & \\ \cline{1-2}
Feni, Khepupara & 1974 & & & \\ \cline{1-2}
\begin{tabular}{@{}c@{}} Kutubdia, Madaripur, \\ Sitakunda, Teknaf
\end{tabular} & 1977 & & & \\ \cline{1-2}
Tangail, & 1987 & & & \\ \cline{1-2}
Chuadanga, Mongla & 1989 & & &\\ \cline{1-2}
Sydpur & 1991 & & & \\ \cline{1-2}
Ambagan(Ctg) & 1999 & & & \\ \hline
\end{tabular}}
\caption{Starting and ending year of data collection for the different stations, total sample size and data type of the dataset}
\label{tab:table1}
\end{center}
\end{table}

\begin{figure}
  \centering
  \includegraphics[scale = .40]{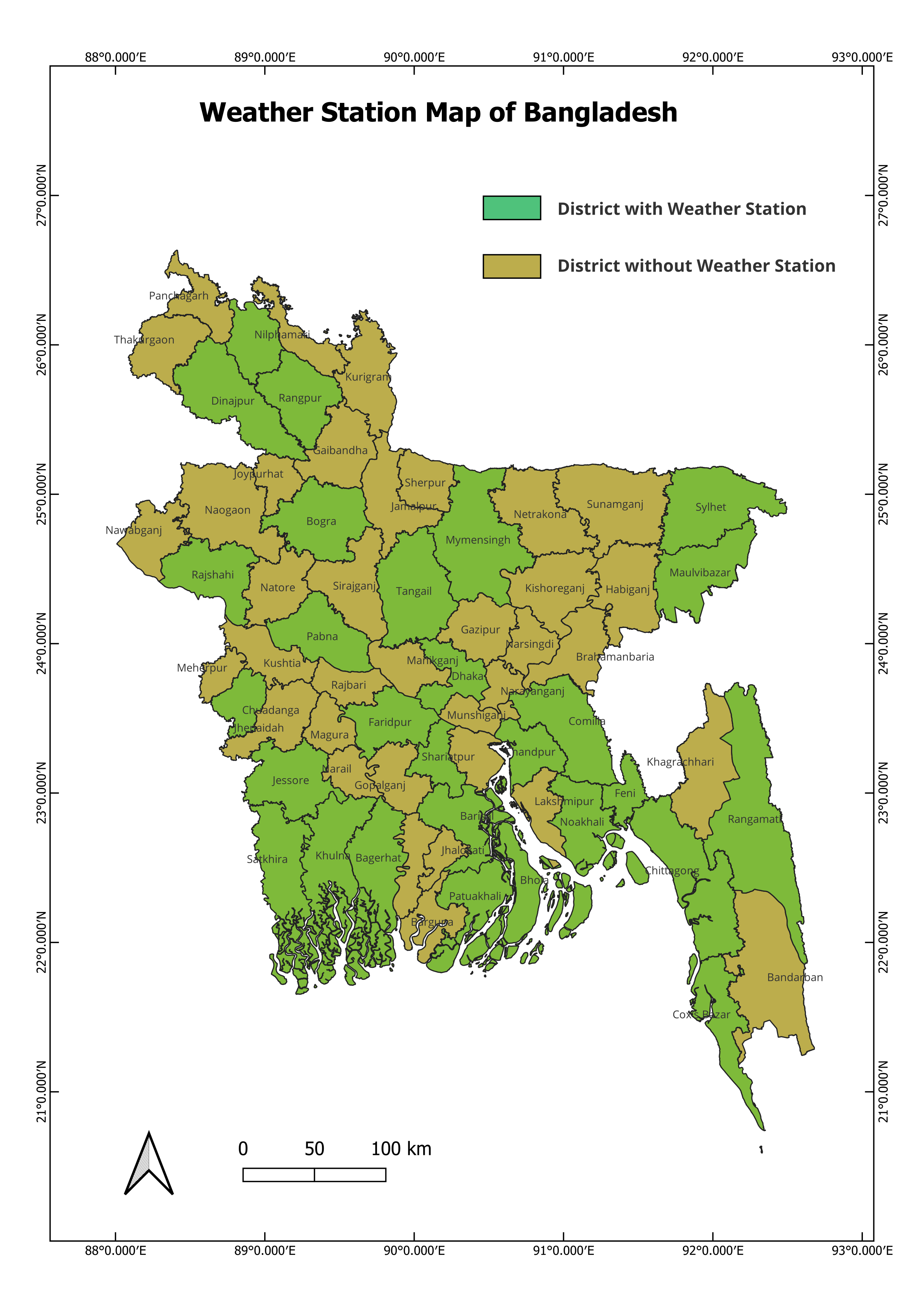}
  \caption{Districts of Bangladesh having weather stations.}
  \label{map_station}
\end{figure}

 In figure \ref{map_station}, we have included a map highlighting the districts of weather stations and used the average daily data.
 Initially, the data was not well organized, and each variable was stored in a different file and format. We have processed the data to make it compatible with the model. The preprocessing steps will be discussed in the upcoming subsections.

\subsection{Insight about Dataset}
\begin{figure*}[]
\centering
\subfigure[]{%
        \includegraphics[width=0.45\linewidth, height=0.3\textwidth]{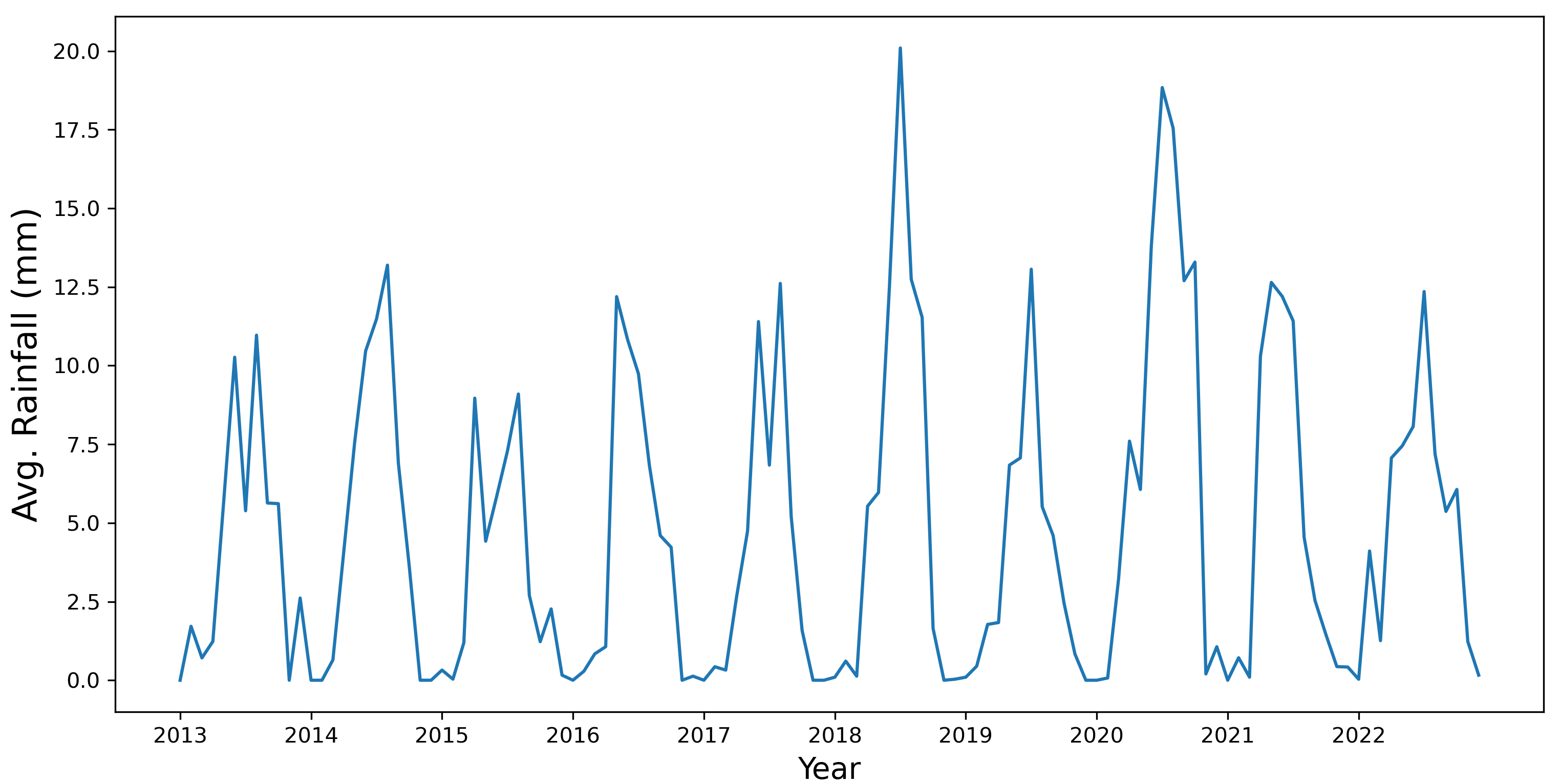}
        \label{rainfall_t}}
\subfigure[]{%{0.3\textwidth}
        \includegraphics[width=0.45\linewidth, height=0.3\textwidth]{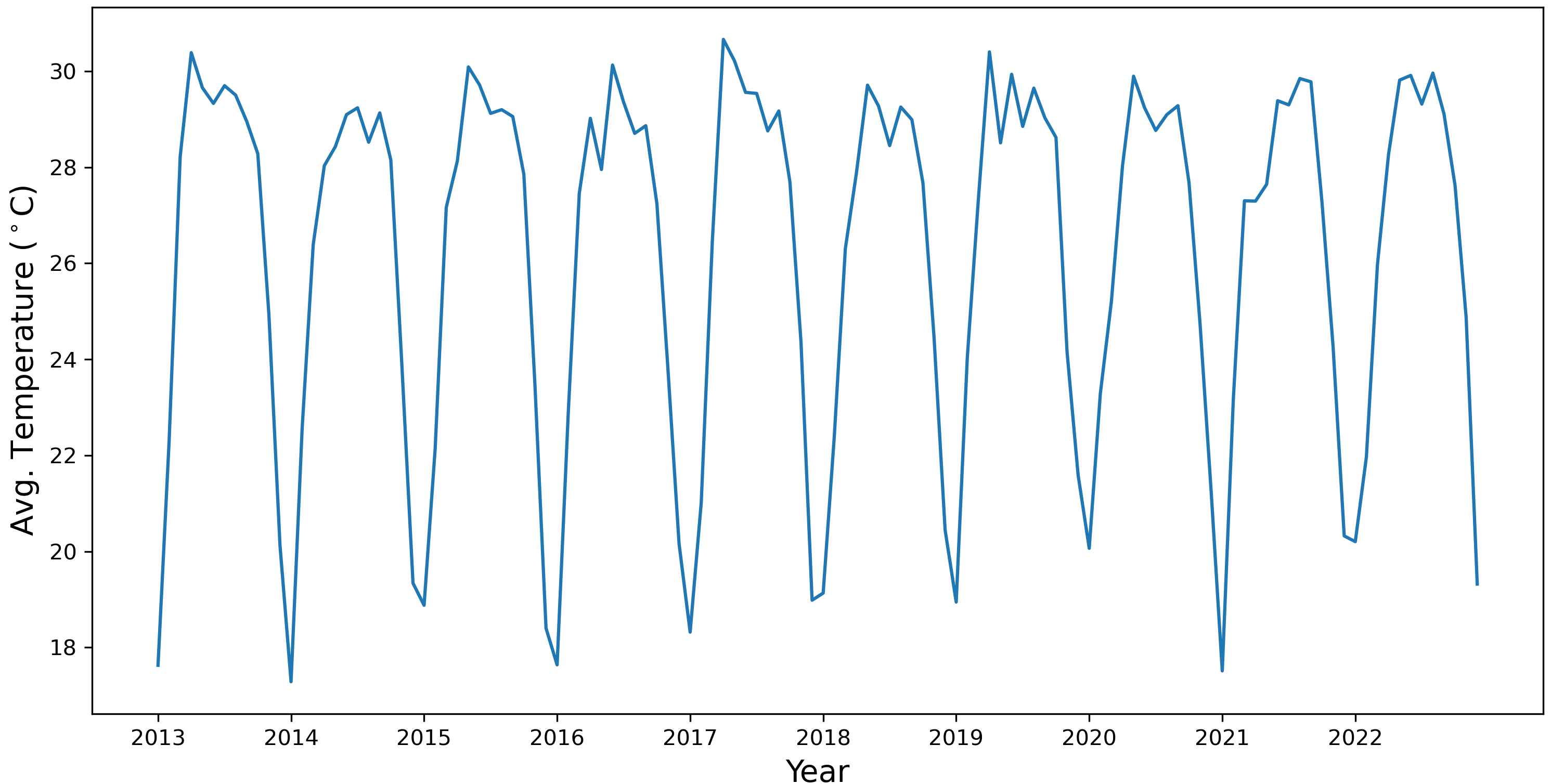}
        \label{temperature_t}}

\subfigure[]{%{0.3\textwidth}
        \includegraphics[width=0.45\linewidth, height=0.3\textwidth]{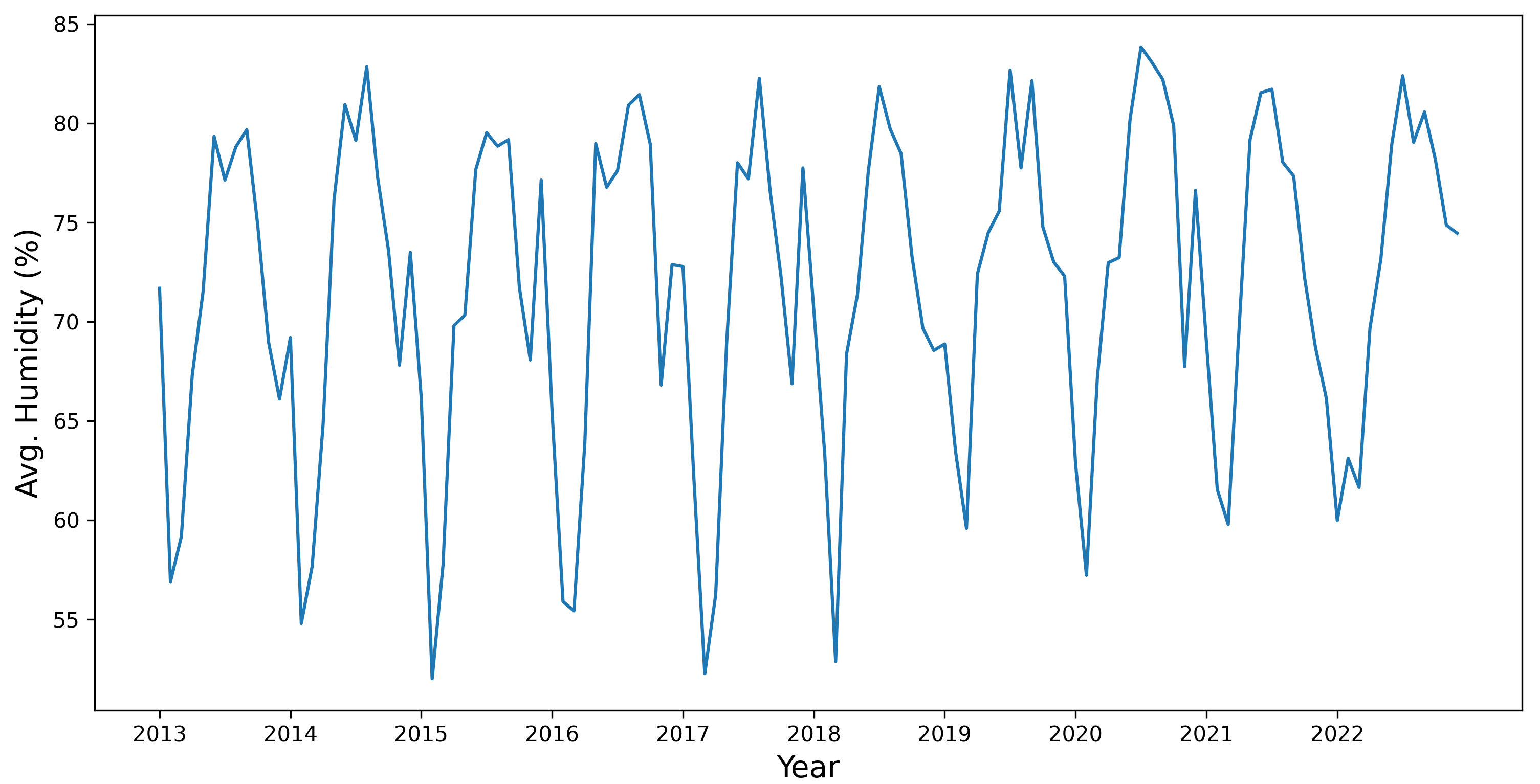}
        \label{humidity_t}}
\subfigure[]{%{0.3\textwidth}
        \includegraphics[width=0.45\linewidth, height=0.3\textwidth]{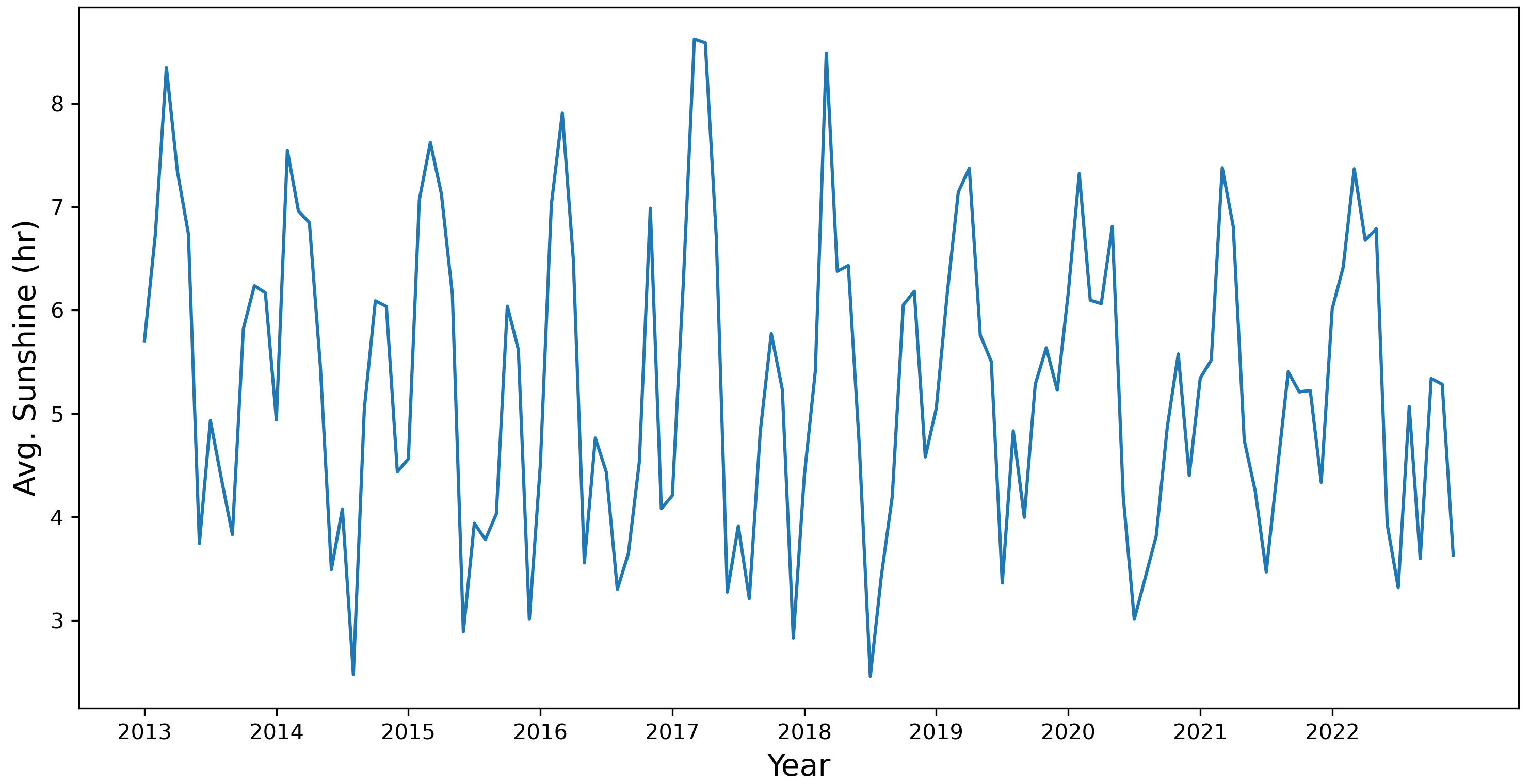}
        \label{sunshine_t}}
\caption{Last 10 years monthly weather trend of average (a)Rainfall, (b)Temperature (c)Humidity and (d)Sunshine of Dhaka City }
\label{trend}
\end{figure*}

Before diving into the details of modeling and forecasting, it's a good idea to get insights into the daily average sunshine, temperature, rainfall, and humidity data. For demonstration purposes, we consider Dhaka City (the capital of Bangladesh). We have plotted the average monthly data of the last 10 years to keep the visualization simple and easily readable.  Figure \ref{trend} shows the weather trend of Dhaka city. 
The time series plots indicate strong and predictable seasonality in average temperature, rainfall, sunshine, and humidity. But we will double-check in the subsection \ref{ADF} to validate whether the data shows seasonality or not. The daily average temperature goes back and forth between high (summer) and low (winter) temperatures. Rainfall, sunshine, and humidity are all seasonal but with less regularity than temperature. However, in terms of both the amplitude and detail of the pattern, the seasonal variations varied substantially across cities.

\subsubsection{Distribution of the Data}

Finding out the data distribution is an important analysis to get an overall summary of the weather features (rainfall, temperature, humidity, and sunshine hour). 
Firstly, we are interested in probability distributions and have selected 7 major cities (Dhaka, Chittagong, Rajshahi, Khulna, Barisal, Comilla, and Sylhet ) in Bangladesh for demonstration purposes. 
\begin{figure*}[]
\centering
\subfigure[]{
  \includegraphics[width=0.45\linewidth, height=0.3\textwidth]{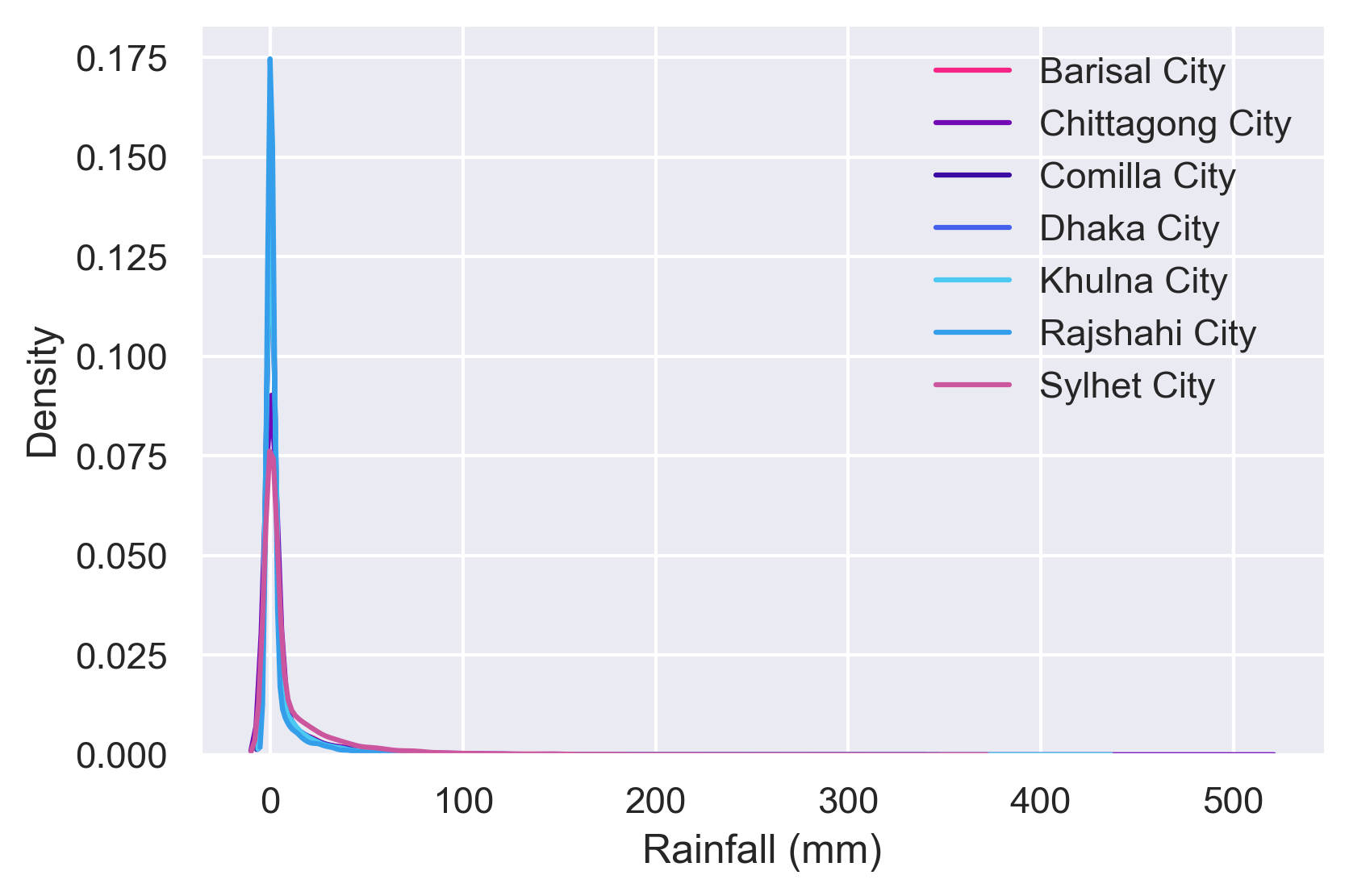}
  \label{rainfall_k}}
\subfigure[]{
  \includegraphics[width=0.45\linewidth, height=0.3\textwidth]{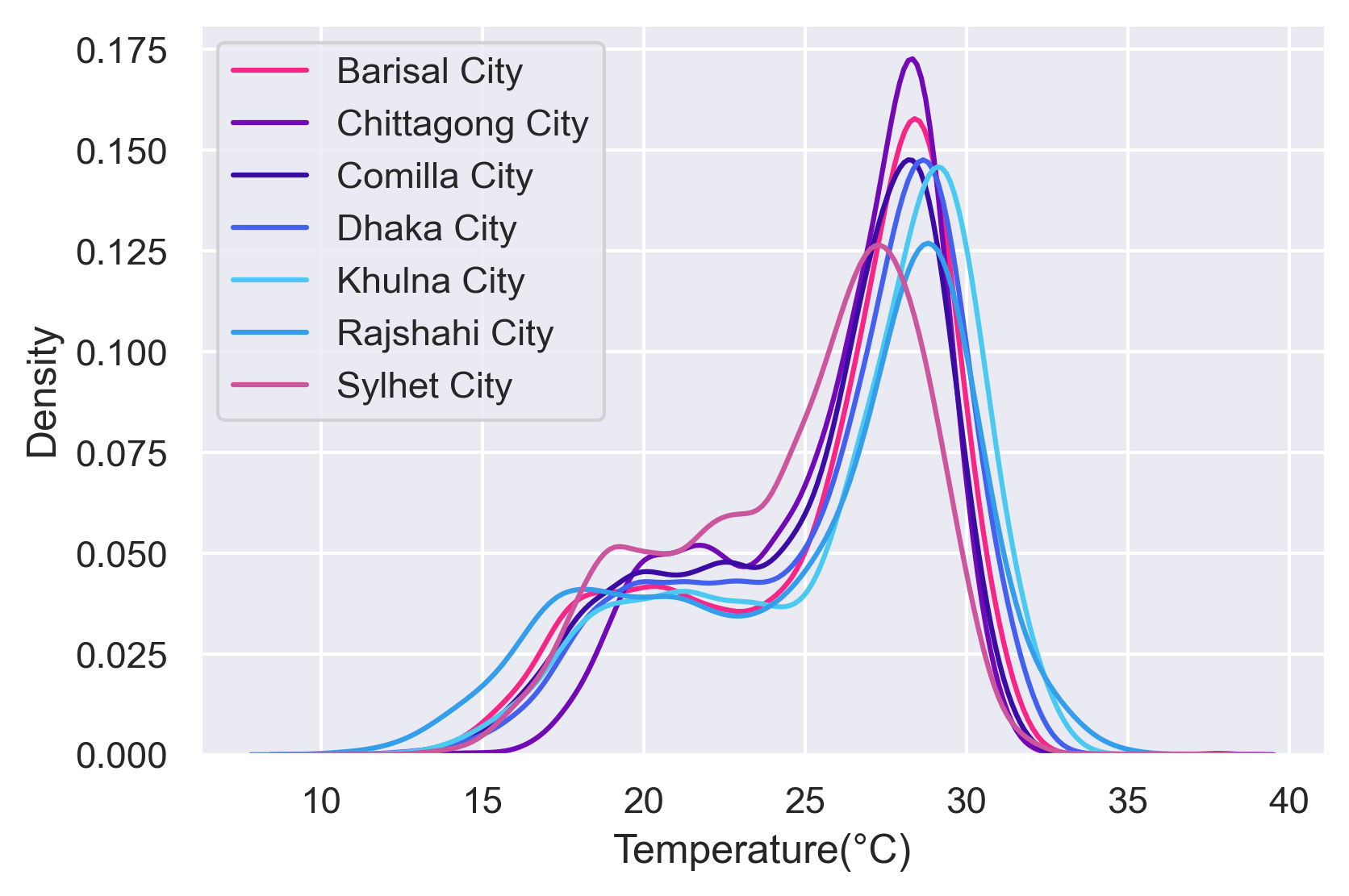}
  \label{temperature_k}}
  
\subfigure[]{
  \includegraphics[width=0.45\linewidth, height=0.3\textwidth]{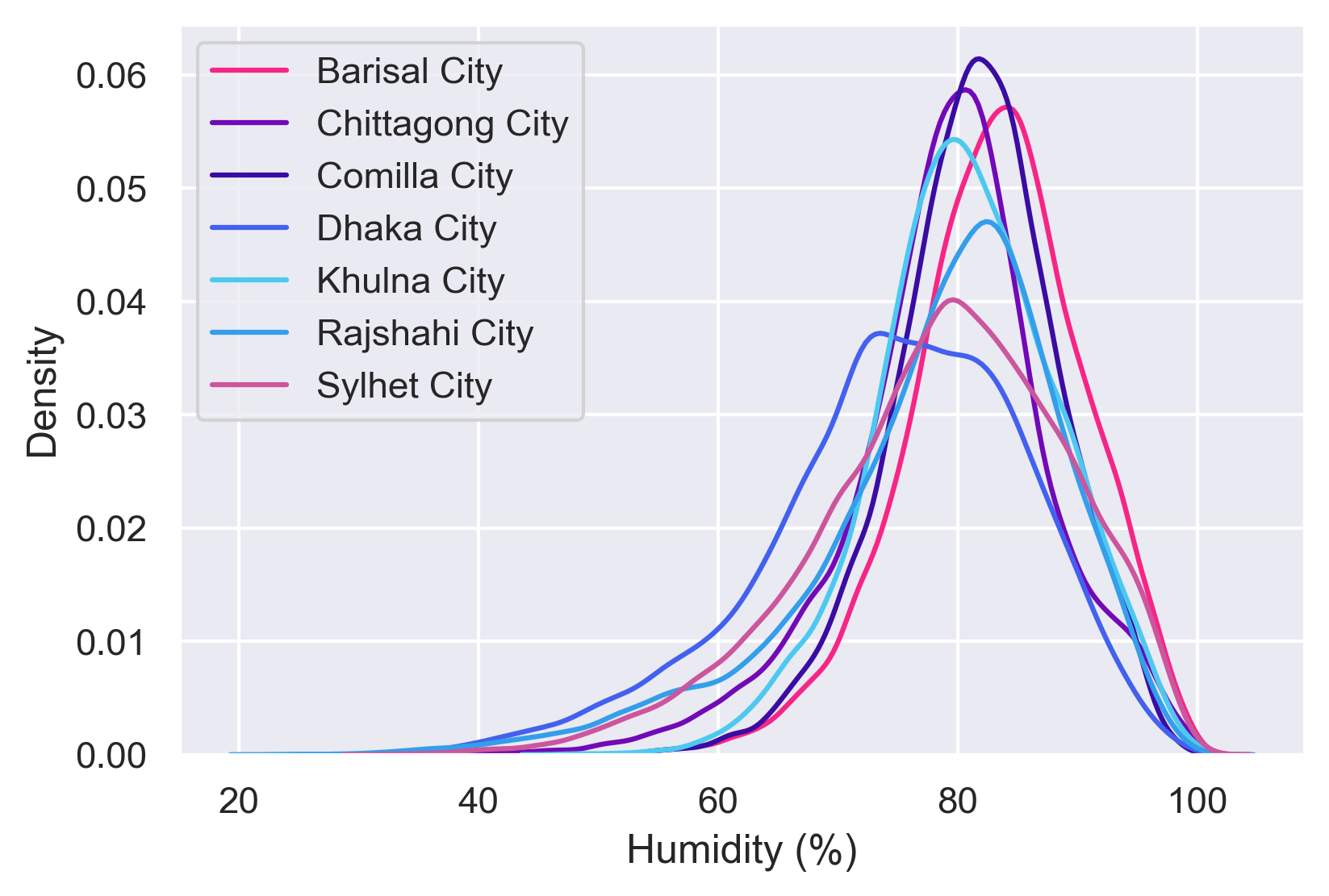}
  \label{humidity_k}}
\subfigure[]{
  \includegraphics[width=0.45\linewidth, height=0.3\textwidth]{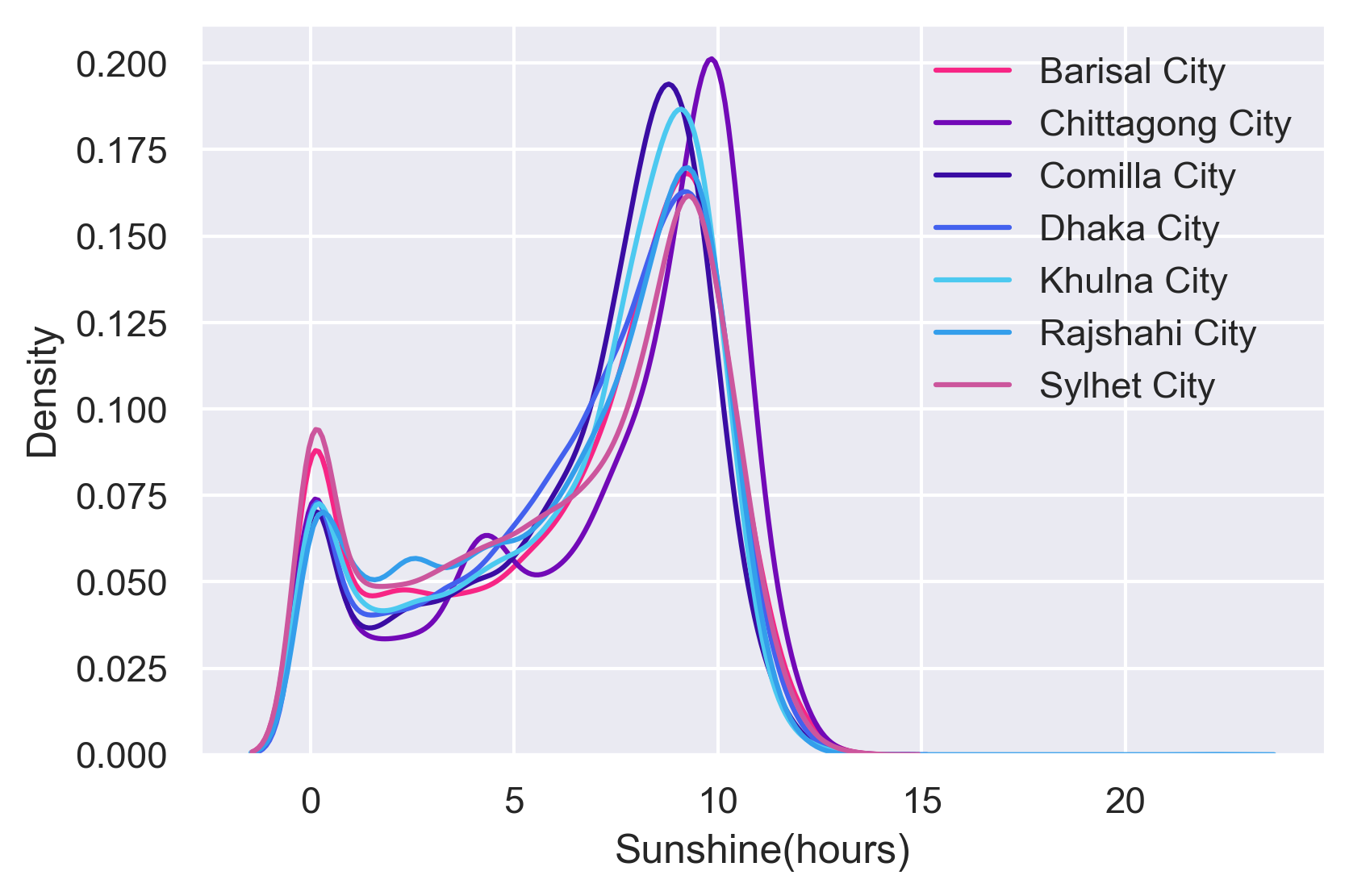}
  \label{sunshine_k}}

\caption{Probability distribution of daily weather records (a) Rainfall, (b)Temperature, (c)Humidity, and (d)Sunshine for 7 major districts of Bangladesh. }
\label{pdf}
\end{figure*}
Figure \ref{pdf} clearly shows the probability distribution, and none of the features is fully normally distributed; rather, they are skewed. Among them, rainfall is highly right-skewed. Most of the data are centered on 0 mm. This indicates most of the days, there is no rainfall, and it ranges from 0 to 500 mm, shown in figure \ref{rainfall_k}. Temperature data is slightly left-skewed; figure \ref{temperature_k} also shows most of the days, the temperature remains in between (25 - 30)$^{\circ}$C. The daily humidity data is nearly normally distributed, and most of the day, we feel a high humidity of 80\%, mentioned in figure \ref{humidity_k}. Typical sunlight duration is 6 to 10 hours, but it may vary between 0 to 15 hours. The plots are identical for the 7 cities. From the data, we can also assume that these graphs project the overall probability distribution of all the weather stations.

\begin{figure}[h!]
  \centering
  \includegraphics[scale = 0.5]{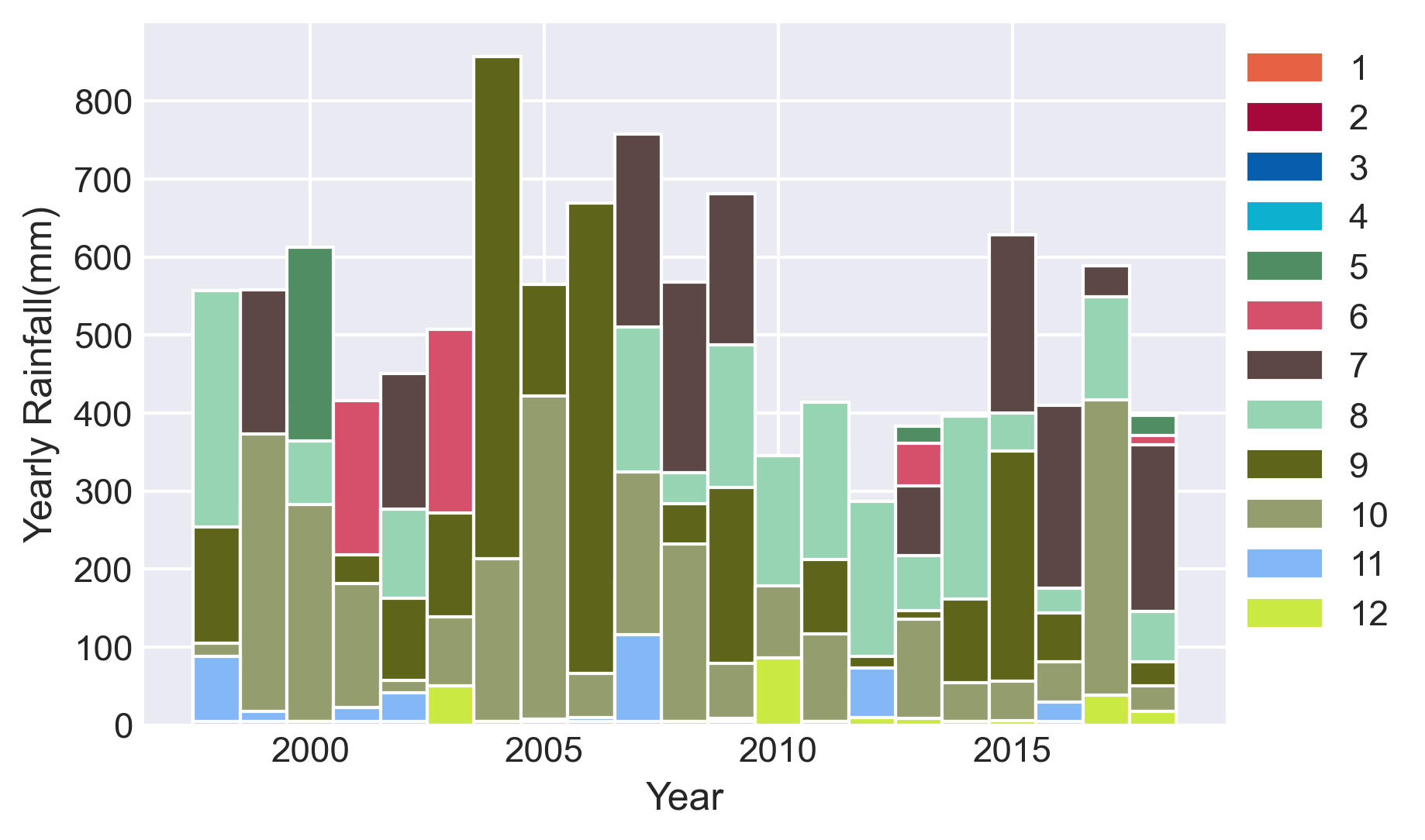}
  \caption{Stacked bar plot for yearly rainfall for the last 20 years of Dhaka city and months are noted with different colors on the right-side legends.}
  \label{rain_bar}
\end{figure}

Secondly, we want to analyze the rainfall and temperature features further because these two features play a vital role in agriculture. In figure \ref{rain_bar}, we plot a stacked bar plot to show the monthly and yearly rainfall of the last 20 years for Dhaka City. Right-side legends indicate the months with different colors.  

\begin{figure}[h!]
  \centering
  \includegraphics[scale = 0.3]{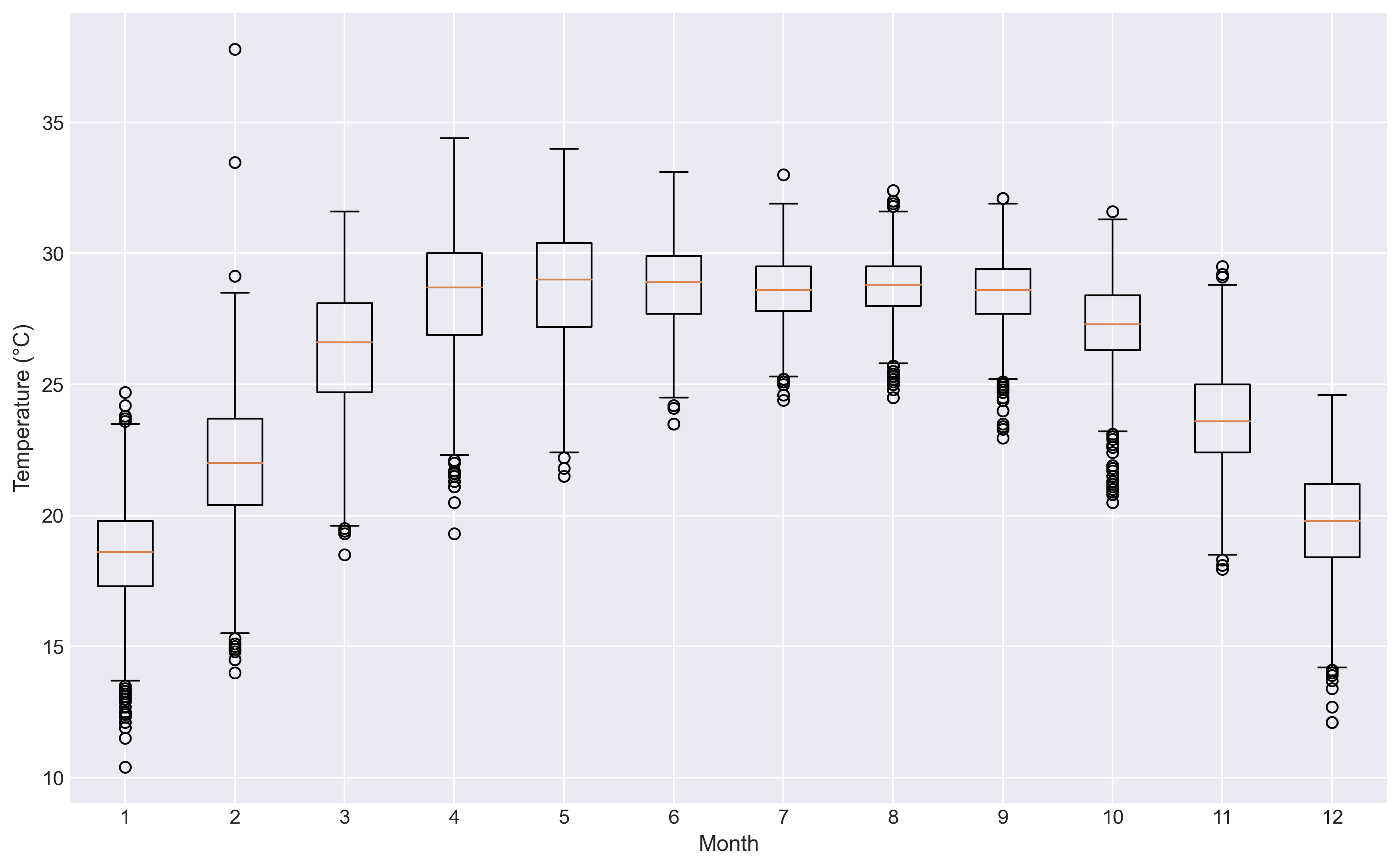}
  \caption{Distribution of monthly temperature of Dhaka city with box plot. }
  \label{temp_box}
\end{figure}

Figure \ref{temp_box} illustrates the variation in temperature throughout the year for Dhaka City. The box plot shows the temperature distribution for each month. Low temperature is recorded for the first and last 3 months of the year. And the temperature becomes high from April to September. 
If we look at the data of other cities, we might get similar results.

\subsubsection{Stationary check using Augmented Dickey-Fuller test}
\label{ADF}
A stationary series is one whose statistical characteristics, such as mean, variance, and covariance, do not change over time or are unaffected by time. In other words, a series without a \textit{trend} or \textit{seasonal} component is said to be stationary in Time Series \cite{nason2006stationary}. It is easier to model a time series when it is stationary. A time series with cyclic behavior is stationary \cite{81Statio79:online}. To be effective, statistical modeling approaches presuppose or demand that the time series will be stationary \cite{Stationa31:online}. We used the Augmented Dickey-Fuller test \cite{mushtaq2011augmented} for the stationary check. We get parameters such as \textit{ADF Statistic-value from the Augmented Dickey-Fuller tests for Critical Values: 5\%, 1\%, and 10\%.} 	
In the context of time series analysis, it can be established that a time series is considered non-stationary if the computed p-value is deemed to be superior to the prescribed significance level of 0.05 and the ADF (Augmented Dickey-Fuller) statistic exceeds the critical values. Conversely, a time series is considered to be stationary if the test statistic is less than the critical value.
Table \ref{tab:table5} shows the stationary check using the Augmented Dickey-Fuller test for Dhaka city. As all the ADF test values are less than the significance level of 0.05, the weather dataset is considered to be stationary. 
\begin{table*}[width=0.7\textwidth]
\centering
\caption{\label{tab:table5} Augmented Dickey-Fuller test  to check stationary for Dhaka city}
\begin{tabular*}{\tblwidth}{|C|C|C|C|C|C|}
\hline
\textbf{Feature} &
  \textbf{Test Statistic} &
  \textbf{P-Value} &
  \begin{tabular}[c]{@{}l@{}}\textbf{Critical} \\ \textbf{Value (5\%)} \end{tabular} &
  \begin{tabular}[c]{@{}l@{}}\textbf{Critical} \\ \textbf{Value (10\%)} \end{tabular} &
  \begin{tabular}[c]{@{}l@{}}\textbf{Stationary/}\\ \textbf{Non-Stationary} \end{tabular} \\ \hline
Rainfall  & -12.29 & 7.70e-23 & -2.86 & -2.57 & stationary \\ \hline
Sunshine  & -12.03 & 2.85e-22 & -2.86 & -2.57 & stationary \\ \hline
Humidity    & -11.34 & 1.05e-20 & -2.86 & -2.57 & stationary \\ \hline
Temperature & -15.44 & 2.86e-28 & -2.86 & -2.57 & stationary \\ \hline
$W_x$          & -14.52 & 5.61e-27 & -2.86 & -2.57 & stationary \\ \hline
$W_y$  & -11.74 & 1.27e-21 & -2.86 & -2.57 & stationary \\ \hline

\end{tabular*}
\end{table*}

\subsection{Data preprocessing}
We have preprocessed the dataset in multiple steps, which will be discussed in the following subsections.  
\subsubsection{Data conversion in the desired format}
The dataset was not found in a usable format. As a result, some pre-processing was necessary. The steps for pre-processing are outlined below. Table \ref{tab:table6} shows an abridged form of the initial look of the dataset humidity data. In our dataset, there were five files - humidity file, rainfall file, sunshine file, temperature file, wind direction, and wind speed file.
\begin{table}
\centering
\caption{\label{tab:table6}Initial look of the humidity data file}
\begin{tabular*}{\tblwidth}{|C|C|C|C|C|C|C|C|}
\hline
\textbf{City} & \textbf{Year} & \textbf{Month} & \textbf{1} & \textbf{.} & \textbf{30} & \textbf{31} & \textbf{Avg.} \\ \hline
\textbf{Dhaka} & 1960.0 & 1.0 & 71 & . & 65 & 61 & 67   \\ \hline
\textbf{Dhaka} & 1960.0 & 2.0   & 60 & . &   &   & 58   \\ \hline
\textbf{Dhaka} & 1960.0 & 3.0   & 70 & .  & 65 & 63 & 60   \\ \hline
\end{tabular*}
\end{table}
Two distinct types of problems have been found in the dataset:
\begin{itemize}
  \item The dataset was not structured.
  \item The dataset had missing values which were blank(null) or star(*).
\end{itemize}
So, we converted monthly data to daily. A detailed description of how to transform the data to our desired format is given in the upcoming subsections.

\subsubsection{Wind speed and wind direction in vector}

The wind direction feature was stored in string format and must be converted into numerical data. First, we converted the wind direction to angular degree format. Angles are not useful as model inputs. Converting the wind direction and velocity columns to a wind vector will reduce the computational complexity of the model.
\begin{table}
\centering
\caption{\label{tab:table10}Wind speed and Wind direction after converting in vector}
\begin{tabular*}{\tblwidth}{|C|C|C|C|C|C|}
\hline
\textbf{Station} & \textbf{Year} & \textbf{Month} & \textbf{Date} & \bf $W_x$ & \bf $W_y$   \\ \hline
Dhaka   & 1960 & 1     & 1    & -11.03 & 2.19 \\ \hline
Dhaka   & 1960 & 1     & 2    & -11.03 & 2.19 \\ \hline
Dhaka   & 1960 & 1     & 3    & -11.03 & 2.19 \\ \hline
Dhaka   & 1960 & 1     & 4    & -11.03 & 2.19 \\ \hline
Dhaka   & 1960 & 1     & 5    & -11.03 & 2.19 \\ \hline
\end{tabular*}
\end{table}
Table \ref{tab:table10} represents the wind speed and direction data after vector conversion.

\subsubsection{Missing value handling}
\begin{figure}[h!]
  \centering
  \includegraphics[scale = 0.5]{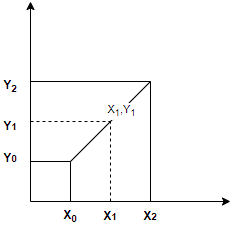}
  \caption{linear interpolation}
  \label{fig:linear1}
\end{figure}

There were multiple missing values in the file. We imputed the missing values. Among several missing value handling techniques for time series data. We have applied the following to handle the missing values. 
\newline
\textbf{Mean imputation:} In this process, we fill the missing values by substituting them with the mean. One downside is that the imputation of every missing value with the mean can distort the seasonality and overlook the data's inherent nature.
\newline
\textbf{Last observation carried forward or backward:} We impute the missing value with the previous or posterior values. This method is unsuitable as it shows problems similar to the mean imputation. 
\newline
\textbf{Linear interpolation:} In linear interpolation, we draw a straight line joining the previous and next points of the missing values as shown in Fig \ref{fig:linear1}. With this method, seasonality is maintained, and no data is distorted. Imputed values are mostly accurate according to the data \cite{linear_interpolation}.
If the coordinates of two known points are (x\textsubscript{0},y\textsubscript{0}) and (x\textsubscript{2},y\textsubscript{2}), we can find the value between the two points with the straight line connecting these points. The equation \ref{slope} of slopes gives the value  of y\textsubscript{2} along the straight line for a value x\textsubscript{2} in the interval (x\textsubscript{0},x\textsubscript{2}).
\begin{equation}
\label{slope}
    \frac{y_{2}-y_{0}}{x_{2}-x_{0}} = \frac{y_{1}-y_{0}}{x_{1}-x_{0}}
\end{equation}
\textbf{Seasonal interpolation:} We address the missing values with the average of corresponding data points of the preceding and succeeding data. In this case, we used linear interpolation because it preserves the seasonality and trend. Any (*) value is replaced with "None," and linear interpolation is further used to impute those "None".

\subsubsection{Merge the files}
We had five files with six parameters - Rainfall, Sunshine, Humidity, Temperature, Wind direction, and Wind speed (wind direction and wind speed are in the same file). We combined all the data into one file. Table \ref{tab:table123} shows a segment of the final view of the dataset.  

\begin{table}
\centering
\caption{\label{tab:table123}Current view of the data from the dataset (Dhaka station)}
\begin{tabular*}{\tblwidth}{|C|C|C|C|C|C|C|}
\hline
\textbf{Year} & \textbf{Month} & \textbf{Date} & \textbf{Rainfall} & \textbf{Sunshine} & \textbf{Humidity} & \textbf{Temp.} \\ \hline
1961 & 1   & 1 & 0.5 & 0.4  & 80 & 21.5 \\ \hline
1961 & 1  & 2  & 1 & 5.9 & 76 & 22.2   \\ \hline
1961 & 1  & 3  & 0 & 4.2 & 75 & 23.1   \\ \hline
1961 & 1  & 4  & 0 & 6.8 & 74 & 19.8    \\ \hline
1961 & 1  & 5  & 0 & 10  & 68 & 16.6    \\ \hline
\end{tabular*}
\end{table}
 As we have 35 stations. We created a single file for each station.

\subsubsection{Normalize the data}
Before training a neural network, it's critical to scale the features. Scaling reduces the complexity of a neural network model. Normalization is a typical method of scaling the data. We perform normalization by subtracting the mean ($\mu$) from each value ($x_i$) and dividing it by the standard deviation ($\sigma$). Equation \ref{eqn:eqn1} can be referred to for this purpose. 
\begin{equation}
	x = \frac{x_i - \mu}{\sigma}
	\label{eqn:eqn1}
\end{equation}
We normalized the dataset by applying equation \ref{eqn:eqn1}.

\subsubsection{Train and test set generation}
The dataset is divided into training, validation, and test sets using a split ratio of (70\%, 20\%, and 10\%). It is noteworthy that the data is not subjected to random shuffling prior to the split for two reasons. Firstly, this approach maintains the continuity of the data and allows for the creation of windows of consecutive samples. Secondly, evaluating the model on the validation and test sets that follow the training data in the original sequence is expected to produce more realistic results. 

\begin{figure*}
\centering
  \includegraphics[width=\textwidth]{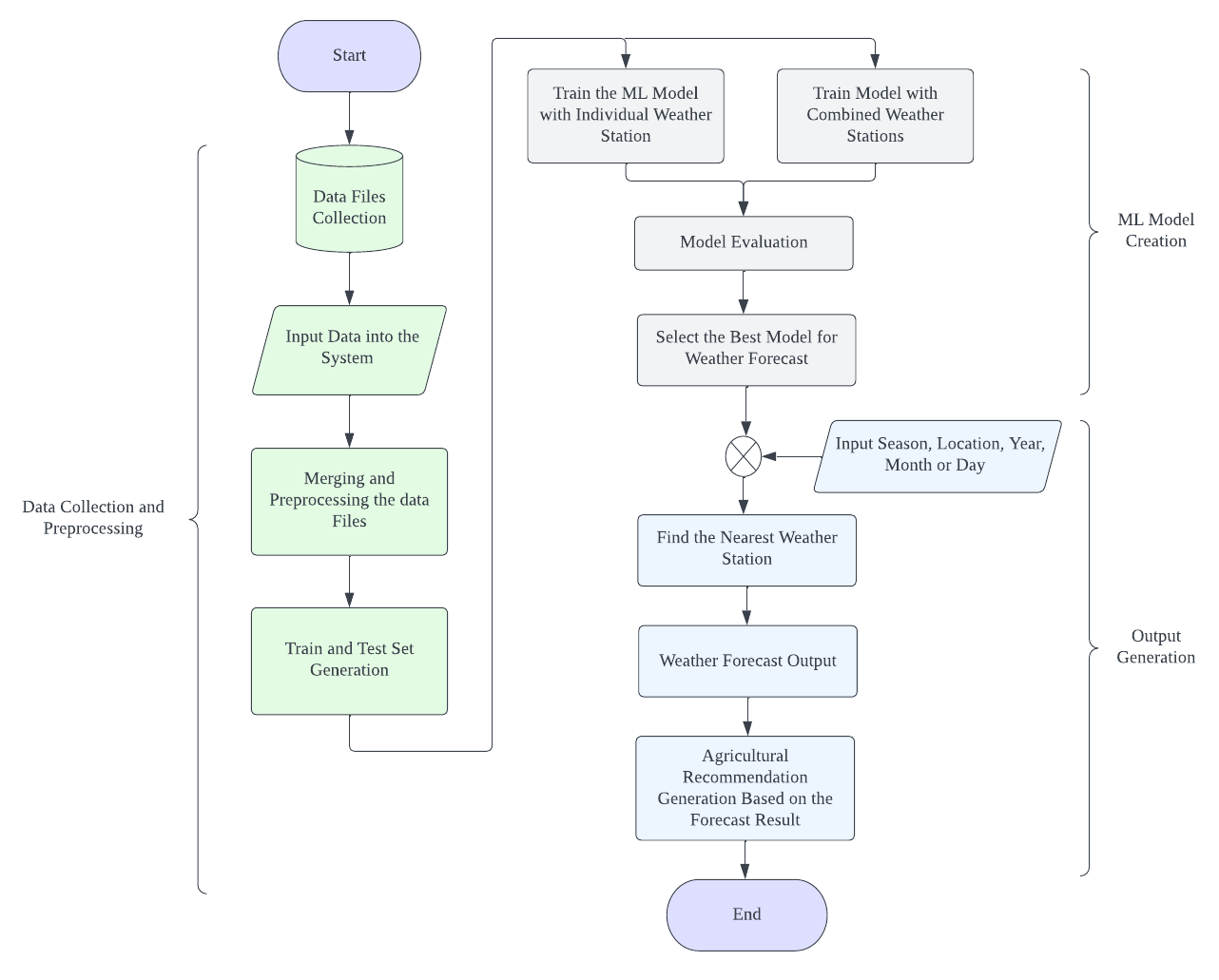}
  \caption{Workflow of our proposed model}
  \label{method}
\end{figure*}

We will discuss our proposed architecture in the next section. Figure \ref{method} represents the entire workflow of our model. 

\section{Proposed Methodology }
\label{proposed_methodology}
This section will outline all the necessary steps for developing our proposed weather forecasting model, encompassing forecast model creation, model training and evaluation, forecast output, and recommendation generation. 

\subsection{Forecast Model creation}
We used a stacked Bi-LSTM model to forecast weather. To create the model, we need to generate a data window, which will be used as a sequence of input for the model. At the same time, the model's hyperparameters will be set.   
\subsubsection{Window generation}
\begin{figure}[]
  \includegraphics[width=\columnwidth]{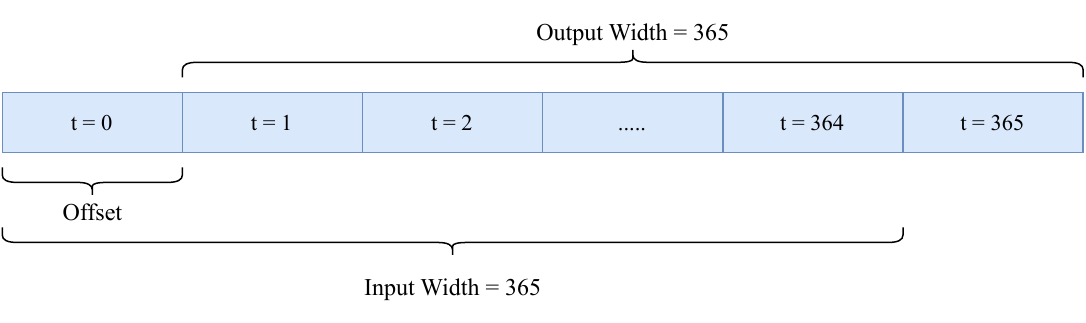}
  \caption{Generating window(Input width=365 ,offset=1,Label width=365)}
  \label{fig:window1}
\end{figure}

\begin{figure}[]
  \includegraphics[width=\columnwidth]{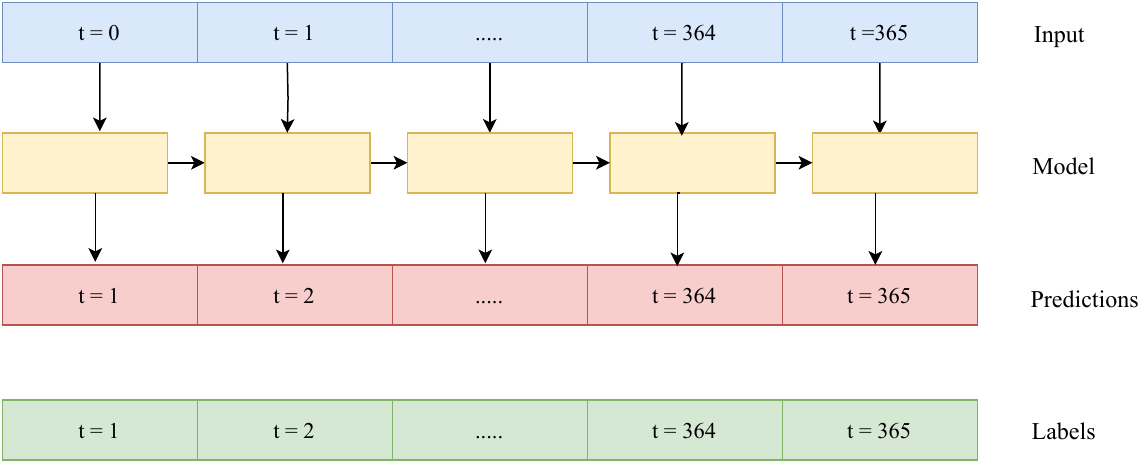}
  \caption{Multi-Timestep Stacked LSTM Architecture for Simultaneous Training}
  \label{fig:window2}
\end{figure}

The application of generated windows in our models aligns with the specified requirements. Our model utilizes a sequence of 365 consecutive data points to forecast weather conditions for the subsequent 365-day period. The construction of the model necessitates obtaining predictions based on consecutive sample windows. The key attributes of these input windows include their width (representing the number of time steps) and the time offset between the input and label windows. 
Figure \ref{fig:window1} clearly depicts how the window is generated for our model. Figure \ref{fig:window2} indicates how generated windows help to train our model.  Selecting features as inputs, labels, or both is crucial in developing the model.

\subsubsection{Customized Bi-LSTM Model Architecture}
\begin{figure*}[hbt!]
\centering
  \includegraphics[width=\textwidth]{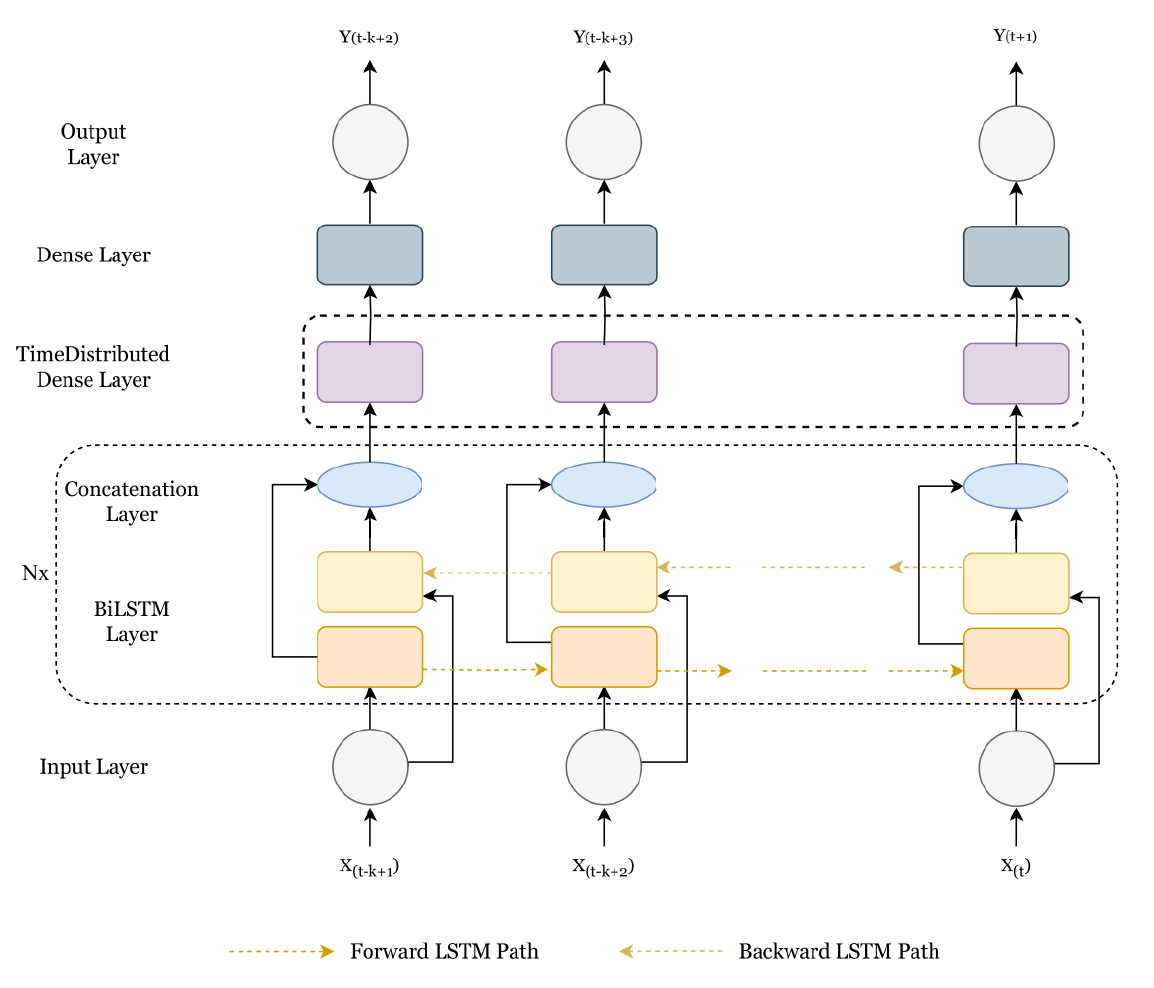}
  \caption{The figure depicts our proposed Stacked Bi-LSTM model with one time distributed layer where $N_x$ is the Bi-LSTM Layer}
  \label{fig:bilstm_fig}
\end{figure*}

We modify the basic Bi-LSTM architecture for higher efficiency. Figure \ref{fig:bilstm_fig} shows the configuration and workflow of our model. In our proposed Bi-LSTM model, Bi-LSTM layers are stacked on top of one another. \textit{The system reduces to a Standard Bi-LSTM model when the number of Bi-LSTM layers is just one ($N_1$), which we used as the baseline.} It defines a Bi-Directional LSTM neural network model with 3 stacked layers. The first layer is a bidirectional LSTM layer with 32 units and a swish activation function, which processes the input sequence in both forward and backward directions and returns the output sequences of all time steps. This output is then passed to the second bidirectional LSTM layer with the same configuration and then to the third one. 
Finally, the output from the last LSTM layer is passed to a TimeDistributed layer followed by a dense layer with 16 units and a swish activation function, which applies the same dense layer to every time step of the input sequence. The output from this layer is then passed to another Dense layer with units equal to the number of output classes to generate the final output. Overall, this model has a total of approximately 169K trainable parameters.
\subsection{Hyperparameter Setting}
The hyperparameter values used for training the model are shown in table \ref{hyperp}. 
\begin{table}
\centering
\caption{\label{hyperp} Hyperparameters for our Bi-LSTM model.}

\begin{tabular*}{\tblwidth}{|C|C|}
\hline
\textbf{Hyperparameter name} & \textbf{Value}           \\ \hline
Input width                  & 365                         \\ \hline
Label width                  & 365                        \\ \hline
Shift                        & 1                        \\ \hline
Window size                  & 365                       \\ \hline
Learning rate                & .001                     \\ \hline
Regularization strength      & L1,L2                    \\ \hline
Activation functions         & swish                    \\ \hline
Number of epochs             & 100                       \\ \hline
Patience                     & 10                       \\ \hline
Loss function                & MAE                      \\ \hline
Metrics                      & MSLE,MSE,SMAPE,R-Squared \\ \hline
Batch size                   & 64                       \\ \hline
Optimizer                    & ADAM                     \\ \hline
\end{tabular*}
\end{table}
The input samples consist of 365 days and predict an output of 365 days, with a one-time step shift between input and output variables. Each sample includes 365 time steps. The learning rate is set to 0.001, and L1 and L2 regularization are applied to the model's weights. The Swish activation function is used, with 100 epochs of training and patience of 10 epochs before early stopping. The models were evaluated based on the Mean Absolute Error (MAE), Mean Squared Error (MSE), Mean Squared Logarithmic Error (MSLE), Coefficient of Determination ($R^2$), and Symmetric Mean Absolute Percentage Error (SMAPE). The Early stopping monitored the validation loss and stopped the training if the same loss occurred 10 times consecutively within 100 epochs. It also restored the model weights from the best value of validation loss. The amount of loss decreases with the increasing epochs. Moreover, the MAE values for other areas are very low in case of training data. It indicates that our proposed model has good training accuracy.

\subsection{Model training}
We trained the model in two different approaches - \textit{i. Individual Bi-LSTM model for each weather station ii. Single Bi-LSTM model for all the weather stations.} Finally, we have selected the outperformed model between the two. 
\subsubsection{Train the ML Model with Individual Weather Station}
Our main goal was to forecast weather and generate agricultural decisions. We trained our model using the Bi-LSTM algorithm with the configuration shown in figure \ref{fig:bilstm_fig}. Here, we took every city's data separately and processed those data. We created the model for every city. The model expects input sequences of shape (365, 4), where  365 is the number of input days, and 4 is the number of features (Rainfall, Temperature, Humidity, and Sunshine) in each time step. We also saved all the training models for 35 stations. We will use the nearest weather station's model to forecast a location's weather parameters.

\subsubsection{Train Model with Combined Weather Stations}
In this case, we combined all the stations in one file. Then we apply one hot encoding for all the stations because it helps to prevent bias in the model since it treats each category equally. One hot encoding is a technique used to convert categorical data into a numerical format that can be used for machine learning algorithms. It works by assigning each category with a unique binary code, where only one bit is on for each code, and all other bits are off. It utilized the same Bi-LSTM architecture used for the previous model but with the input shape of (365, 39) (where 39 is the input feature in each time step). In this model, input features are increased because we used every city as a column using one hot encoding.

\subsection{Model Evaluation}
In the field of machine learning and statistical analysis, the evaluation of predictive models is essential to assess their performance and effectiveness. Several evaluation metrics are commonly employed to measure the accuracy and quality of regression models. In this work, we focus on five widely used metrics: \textit{Mean Absolute Error (MAE), Mean Squared Error (MSE), Mean Squared Logarithmic Error (MSLE), Coefficient of Determination (R\^2), and Symmetric Mean Absolute Percentage Error (SMAPE).}\newline
The \textbf{\textit{Mean Absolute Error}} is a common regression metric that measures the average absolute difference between the predicted and actual values. It provides a measure of the magnitude of errors without considering their direction.
\begin{equation} \label{eqn1}
    MAE = \frac{1}{n} \sum_{i=1}^{n} |y_i - \hat{y}_i| 
\end{equation}
The \textbf{\textit{Mean Squared Error}} is another widely used regression metric that calculates the average of the squared differences between the predicted and actual values. It amplifies larger errors due to the squaring operation.
\begin{equation} \label{eqn2}
    MSE = \frac{1}{n} \sum_{i=1}^{n} (y_i - \hat{y}_i)^2 
\end{equation}
The \textbf{\textit{Mean Squared Logarithmic Error}} is often used for tasks where the predicted values are expected to have a wide range. It measures the average of the squared logarithmic differences between the predicted and actual values.
\begin{equation} \label{eqn3}
    MSLE = \frac{1}{n} \sum_{i=1}^{n} (\log(1 + y_i) - \log(1 + \hat{y}_i))^2
\end{equation}
The \textbf{\textit{Coefficient of Determination}}, often denoted as $R^2$, represents the proportion of the variance in the dependent variable that can be explained by the independent variables. It indicates the goodness of fit of the regression model.
 \begin{equation} \label{eqn4}
     R^2 = 1 - \frac{\sum_{i=1}^{n} (y_i - \hat{y}i)^2}{\sum{i=1}^{n} (y_i - \bar{y})^2}
 \end{equation}
The \textbf{\textit{Symmetric Mean Absolute Percentage Error}} is an alternative metric for evaluating the accuracy of forecasts or predictions. It calculates the average percentage difference between the predicted and actual values, considering the magnitude of the values.
\begin{equation} \label{eqn5}
     SMAPE = \frac{100}{n} \sum_{i=1}^{n} \frac{|y_i - \hat{y}_i|}{(|y_i| + |\hat{y}_i|)/2}
\end{equation}
In the above equations \ref{eqn1},\ref{eqn2},\ref{eqn3},\ref{eqn4} and \ref{eqn5}, $y_i$, $\hat{y}_i$ and $n$ are consecutively the actual value, predictive value, and the total number of predicted values.

\subsection{Find the nearest Weather Station}
\label{nearest}
We have only 35 weather stations in Bangladesh which don't cover every district and sub-district of the country. In this situation, we need a way to find out the nearest weather station from any location in Bangladesh. To do so, we have used the Haversine formula. 
The Haversine formula takes the longitude and latitude of any two locations and calculates the distance between the locations \cite{chopde2013landmark}. It returns the value in Kilometres and feet. It helps to calculate the distance between two great circles, which defines the shortest distance over the earth's surface.
\begin{equation}
\label{eu_eqn}
\resizebox{.9\hsize}{!}{${D=\ 2Rsin^{-1}(\sqrt{sin^2(\frac{\Theta_2-\Theta_1}{2})+cos(\Theta_1)cos(\Theta_2)sin^2(\frac{(\Psi_2-\Psi_1}{2}}))}$}
\end{equation}
where, $D$ is the distance between two locations, $R$ is the radius of earth(6371 km), $\Theta_1$ and $\Theta_2$ represents the latitudes of two locations, $\Psi_1$ and $\Psi_2$ represents the longitudes of two locations.
In the equation \ref{eu_eqn}, we pass the input location and all the weather station's coordinates (longitude and latitude) to calculate the distance of the input location from all the weather stations. Finally, we select the nearest station based on the shortest distance.

\subsection{Agricultural Recommendations and Weather Forecast}
\label{recommendation} 
Agriculture is highly dependent on natural phenomena, and weather plays a key role in farming. Therefore, future weather predictions and suggestions by analyzing weather forecast output might be helpful for users. Due to the lack of proper education and knowledge, farmers in developing countries like Bangladesh are not conscious of the effect of the weather on farming \cite{ali2020farmer}. Our proposed model will generate an overall suggestion based on future weather to assist farming. We have used Bangladeshi cropping season and demographic information for implementation purposes.  

\begin{table}
   
    \centering
     \caption{\label{season1}Cropping season of Bangladesh}
    \begin{tabular*}{\tblwidth}{|L|L|L|}
     \hline
    \textbf{Season} & \textbf{Sub-season}  & \textbf{Duration}   \\ \hline 
     Rabi                 & -              & Mid-November to Mid-March        \\ \hline
     \multirow{2}{4em}{Kharif}  & Kharif-1 & Mid-March to Mid-July \\ \cline{2-2} \cline{3-2}
    & Kharif-2 & Mid-July to Mid-November \\ \hline
   
    \end{tabular*}
\end{table}

Mainly, Bangladesh has two cropping seasons - \textit{Rabi and Kharif}. The Rabi crops are known as the winter crops grown from mid-November to mid-March. Kharif crops are grown in the summer or monsoon weather (mid-March to mid-November). Kharif season is further classified into two sub-divisions Kharif-1 (mid-March to mid-July) and Kharif-2 (mid-July to mid-November) \cite{hossain2018climate}. Table \ref{season1} represents information regarding the cropping seasons of Bangladesh. 

In the upcoming subsections, We will describe \textit{i) how the model will predict future weather, and ii) generate recommendations based on the output weather.} 

\begin{figure*}
    \centering
    \includegraphics[scale=0.75]{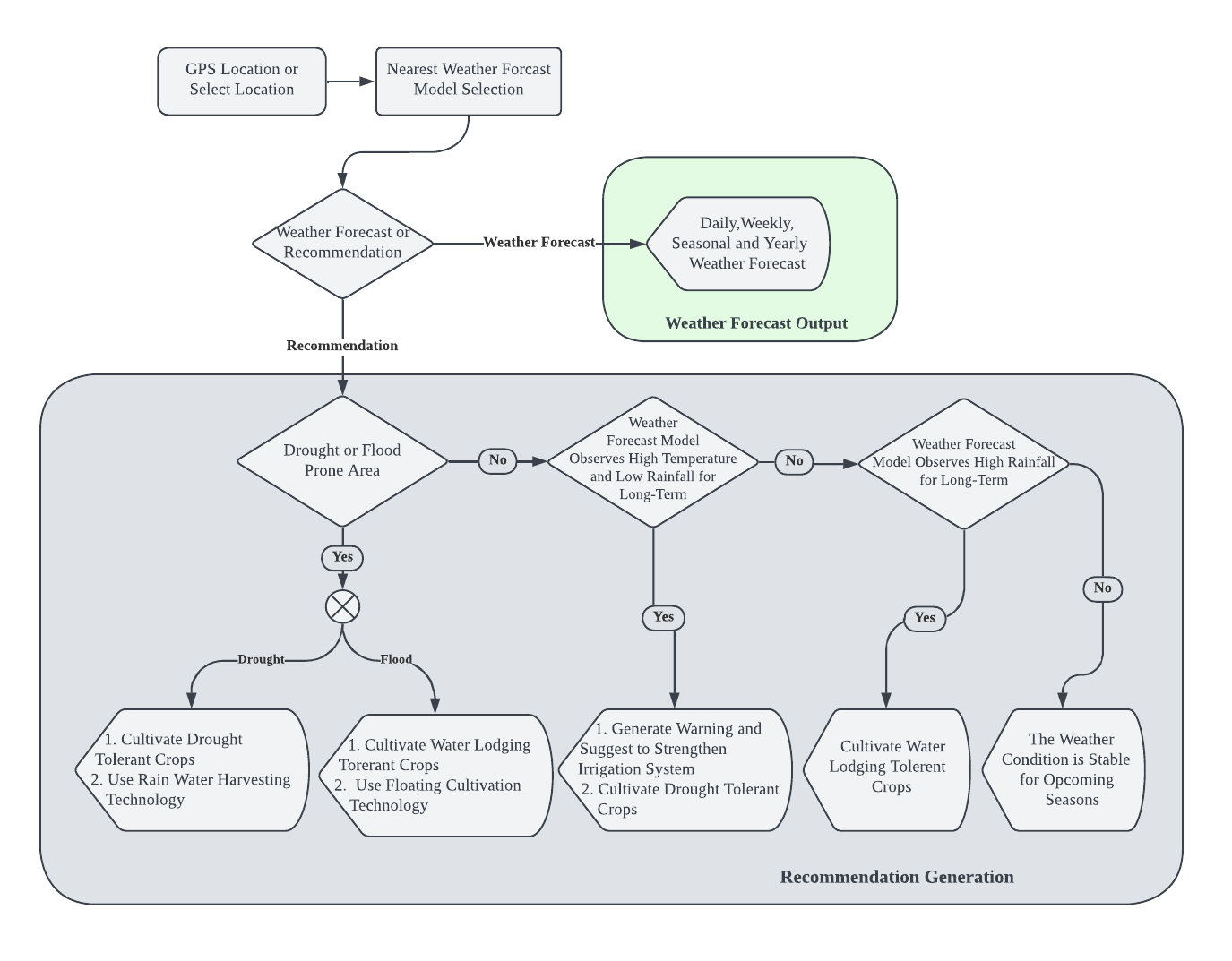}
    \caption{Location-Specific Agricultural Recommendation generation based on Weather-Forecast Output and Knowledge-Base System. }
    \label{recommendation}
    \end{figure*}

\subsubsection{Weather Forecast Output}
Previously, we have trained 35 Bi-LSTM models to forecast the weather of the entire Bangladesh. Our prediction model is ready to forecast weather and can automatically track the user's location, or the user can select a specific location. In the next step, the model will select the nearest weather station's forecast model to predict the weather of the selected location, according to the process described in the subsection \ref{nearest}. Our model can predict daily, weekly, monthly, seasonal \textit{(according to cropping season as shown in table \ref{season1})}, and yearly weather \textit{(Temperature, Rainfall, Humidity, Sunshine)}. The process of weather forecasting is clearly depicted in figure \ref{recommendation}. It will help the users to make appropriate farming decisions by considering future weather.  

\subsubsection{Agricultural Recommendation}
Bangladesh is a small country, having an area of 147,630 km\textsuperscript{2} \cite{Bangladesh}. Though the country is small, it has diversified climates, and it is changing day by day. There are also some natural phenomena like floods, drought, etc. Basically, geographical location is responsible for natural hazards.  So, agricultural suggestions based on geographical location and weather forecast are needed. 
\begin{figure}
    \centering
    \includegraphics[scale=0.4]{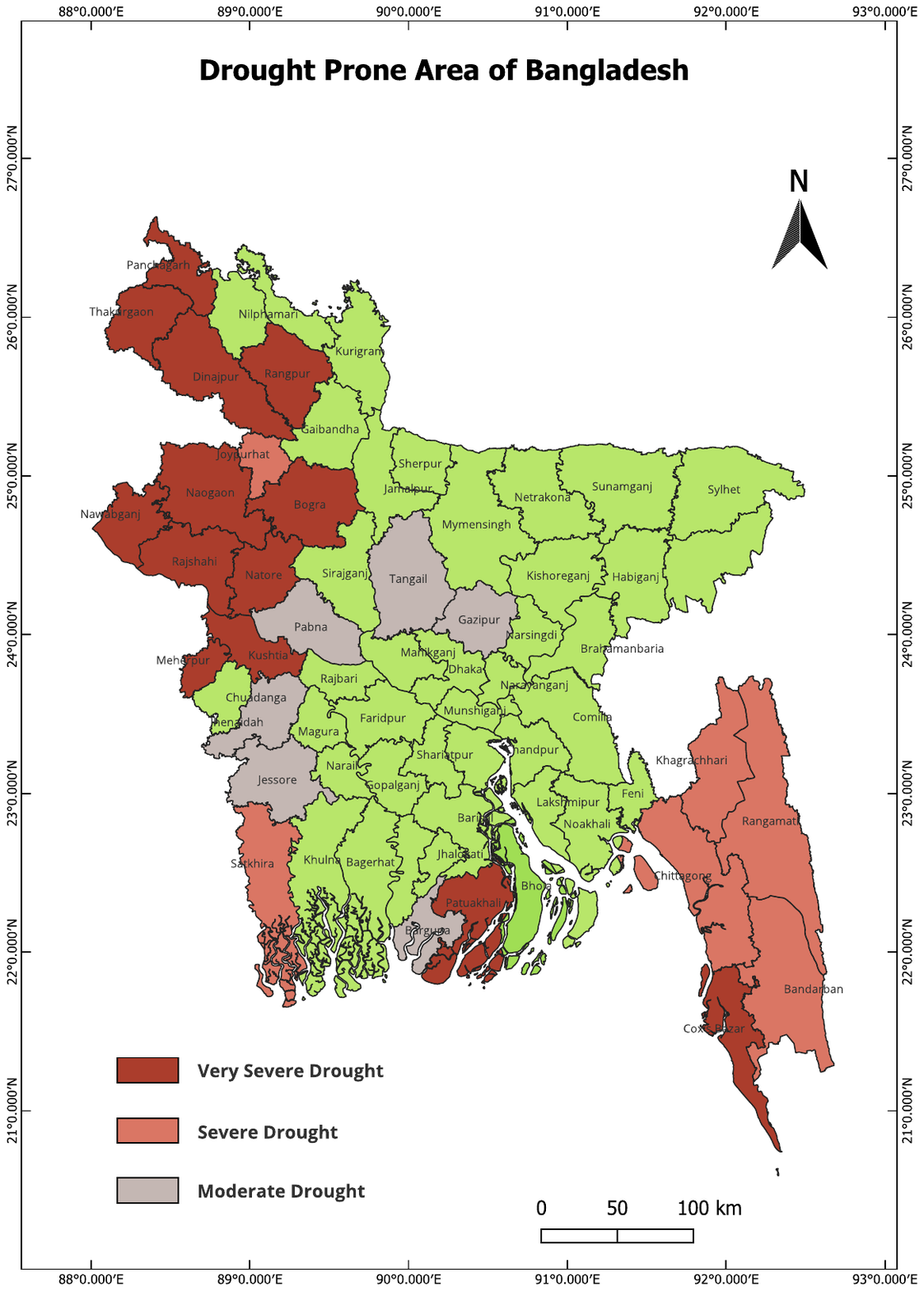}
    \caption{Drought-prone districts of Bangladesh \cite{drought}, and these districts are considered as drought-prone regions in our decision system.}
    \label{drought}
\end{figure}
Recommendations are generated as follows- 
\begin{itemize}

 \item This proposed model considers low-lying (flood-prone) and drought-prone areas. The northwest part of Bangl- adesh is the most drought-prone area. Some other districts are also drought-prone for demographic location \cite{drought}. The map (Figure \ref{drought}) shows the very severe, severe, and moderate drought-prone districts of Bangladesh. Additionally,  many rivers and low-lying lands flood away every year in Bangladesh, especially in the rainy season. We have used information regarding low-lying regions of Bangladesh published by Bangladesh Agricultural Research Council(BARC) \cite{floodprone}.
 
\begin{table*}[width=0.9\textwidth]
\centering
\caption{\label{crop_database} Season-based crops for drought and flood-prone areas of Bangladesh \cite{Divers, nasim2017distribution, islam2012crop, hajong2020study}.}
\begin{tabular*}{\tblwidth}{|C|C|C|}
\hline
\textbf{Season} &  \textbf{Droght Tolerant} &
  \textbf{Flood Lodging Tolerant} \\ \hline

 \textbf{Rabi}  & 
 \begin{tabular}{@{}c@{}c@{}} T. aman rice, Mustard, Wheat, Grasspea Mungbean, \\ Watermelon, Cucumber, Sugarcane, Tomato, Okra, \\ Different vegetables \end{tabular} & 
 \begin{tabular}{@{}c@{}c@{}}Potato, Maize, Mustard, Chickpea, \\  Onion, Tobacco, Groundnut, Ropa aman, \\ Chili, Pulses, Different vegetables \end{tabular} \\  \hline 

\textbf{Kharif-1} & 
\begin{tabular}{@{}c@{}}Jute, Sesame, Groundnut, Maize, \\ Different Vegetables \end{tabular} & \begin{tabular}{@{}c@{}}Maize, Jute, Boro rice, Soybean, \\ Groundnut, Chili, Different vegetables \end{tabular} \\ \hline 

\textbf{Kharif-2} & 
\begin{tabular}{@{}c@{}}Sunflower, Soybean, Betel leaf, Chilli, \\ Different Vegetables \end{tabular} & \begin{tabular}{@{}c@{}}Maize, Sunflower, Soybean, Aus rice, \\ Chili, Different vegetables \end{tabular} \\ \hline

\end{tabular*}
\end{table*}

According to the flowchart \ref{recommendation}, if the user selects the recommendation option, our model will first check whether it is a drought or flood-prone area or not. If the area is listed in the flood or drought-prone database, our model will suggest some suitable crops according to the table \ref{crop_database} and modern technology for hazard-prone areas.  So, the farmers, especially inexperienced farmers, will be benefited by minimizing the loss and maximizing the profit. 

\begin{table}
    
    \centering
     \caption{\label{threshold}Avg. monthly max temperature and rainfall of Bangladesh \cite{weather}, and these values will be used as a threshold for decision-making.}
    \begin{tabular*}{\tblwidth}{|C|C|C|}
     
    \hline
     \textbf{Month} & \textbf{Max. Temperature (\textdegree{C})}  & \textbf{Rainfall(mm)}\\
     \hline 
     \textbf{January} & 25.3 & 10.1 \\
     \hline
     \textbf{February} & 28.1 & 20.2 \\
     \hline
     \textbf{March} & 32.1 & 41.6 \\
     \hline
     \textbf{April} & 33.4 & 118.6 \\
     \hline
     \textbf{May} & 33 & 227.8 \\
     \hline
     \textbf{June} & 32 & 367.3 \\
     \hline
     \textbf{July} & 31.4 & 474.4 \\
     \hline
     \textbf{August} & 31.6 & 384.7 \\
     \hline
     \textbf{September} & 31.8 & 296.7 \\
     \hline
     \textbf{October} & 31.4 & 189.4 \\
     \hline
     \textbf{November} & 29.4 & 29.4 \\
     \hline
     \textbf{December} & 26.2 & 5.3 \\
     \hline
    \end{tabular*}
    
    \end{table}
    
\item The weather pattern is continuously changing, and many regions of Bangladesh are being affected by heat waves and over-rainfall. We have considered some thresholds, as shown in table  \ref{threshold}, to identify the weather condition. Table 
 \ref{threshold} contains the maximum temperature and total average rainfall (mm) for each month in Bangladesh. If our proposed model finds the forecast temperature is significantly high or the rainfall is significantly low compared to the threshold values over the year of a location, it will generate a warning and suggest drought-tolerant crops for cultivation. Furthermore, if the forecast model predicts long-term heavy rainfall, it will suggest water-lodging tolerant crops. The process is graphically represented in the figure \ref{recommendation}.

\end{itemize}

In short, our proposed model generates recommendations based on the weather forecast model's output and some knowledge-based systems. These agricultural recommendations will help the farmers by raising awareness about weather conditions, which will help them to make appropriate decisions in farming. Apart from weather forecasts, the model can generate some suitable crop suggestions in the flood and drought-prone areas of Bangladesh. \textit{Our model is not limited to work only for Bangladesh. It can be applied to any country over the world. We only need to train the model with the weather data of that country or region, keeping the process identical.}

\section{Evaluation and Experimental Results}
\label{evaluation}
Without proper evaluation, we can't justify a model's acceptability and robustness. To do so, we need to apply appropriate evaluation metrics carefully. LSTM is one of the best time series forecasting models \cite{LINDEMANN2021650}. So, We have compared our proposed model with other LSTM models from different perspectives. We created different forecast models with standard LSTM, Bi-LSTM, stacked Bi-LSTM with various layers, and so on. We compared all the models based on efficiency and picked the best one. We also demonstrate some results of our proposed model. All the descriptions and results are included in the subsections. 

\subsection{Result Analysis}

\subsubsection{Best Model Selection}
We evaluated different LSTM models and selected the best one based on performance. For execution purposes, we used a personal computer for this work. The specifications of the computer are compiled in Table \ref{exp_setup}. 
\begin{table}[!hbt]
\centering
\caption{\label{exp_setup}System Configuration}
\begin{tabular*}{\tblwidth}{|C|C|} 
\hline
\textbf{Component} & \multicolumn{1}{c|}{\textbf{Specification}}            \\ 
\hline
GPU                & 2x Nvidia Quadro p4000(8GB)                    \\ 
\hline
Processor          & 16x Intel(R) Xeon(R) Bronze 3106 \\ 
\hline
RAM                & 128GB                                          \\ 
\hline
Storage            & 4TB                                            \\ 
\hline
OS                 & Windows Server 2016(64 bit)~                   \\
\hline
\end{tabular*}
\end{table}
As there are 35 stations, firstly, we created standard LSTM and Bi-LSTM models for each station, i.e., 35 models for 35 stations, respectively. Furthermore, we created standard LSTM and Bi-LSTM models combining all the stations. In the combined model, we used one-hot encoding for the 35 stations. 
 Primarily, the models' input shapes were (365,4), where 365 is the window size with 4 features.  The features were temperature, rainfall, humidity, and sunshine. In the case of the combined model, both for LSTM and Bi-LSTM, we achieved R\textsuperscript{2} value of 0.0131. It signifies the models are not a good fit for the forecast. For the standard LSTM model with individual stations, the average R\textsuperscript{2} value is 0.413, which is better than the combined model but still below the standard. The standard Bi-LSTM model outperforms by achieving the average R\textsuperscript{2} value of 0.91 comparing the previous architectures. So, we picked the Bi-LSTM model for the proposed forecast model. The experimental results are shown in table \ref{combined}. We further fine-tuned and customized the Bi-LSTM model for better performance. 

\begin{table*}[]
\caption{\label{combined} Results of Combined and Individual station models with the standard LSTM and Bi-LSTM architectures.}
\resizebox{\textwidth}{!}{
\begin{tabular}{|c|c|c|c|}
\hline
\textbf{Stations}             & \textbf{Architecture} & \textbf{Features}                                          & \textbf{Average R\textsuperscript{2} Value} \\ \hline
Combined Stations             & Standard LSTM         & \multirow{3}{*}{Temperature, Rainfall, Humidity, Sunshine} & 0.0131                    \\ \cline{1-2} \cline{4-4} 
Combined Stations             & Standard Bi-LSTM      &            & 0.0131                    \\ \cline{1-2} \cline{4-4} 
Individual Stations (average) & Standard LSTM         &                                                            & 0.413                     \\ \cline{1-2} \cline{4-4}
\textbf{Individual Stations (average)} & \textbf{Standard Bi-LSTM}         &                                                            & \textbf{0.91}                     \\ \hline
\end{tabular}}

\end{table*}

\subsection{Model Training Loss}
\begin{figure*}[]
\centering
\subfigure[]{
  \includegraphics[width=0.45\linewidth, height=0.27\textwidth]{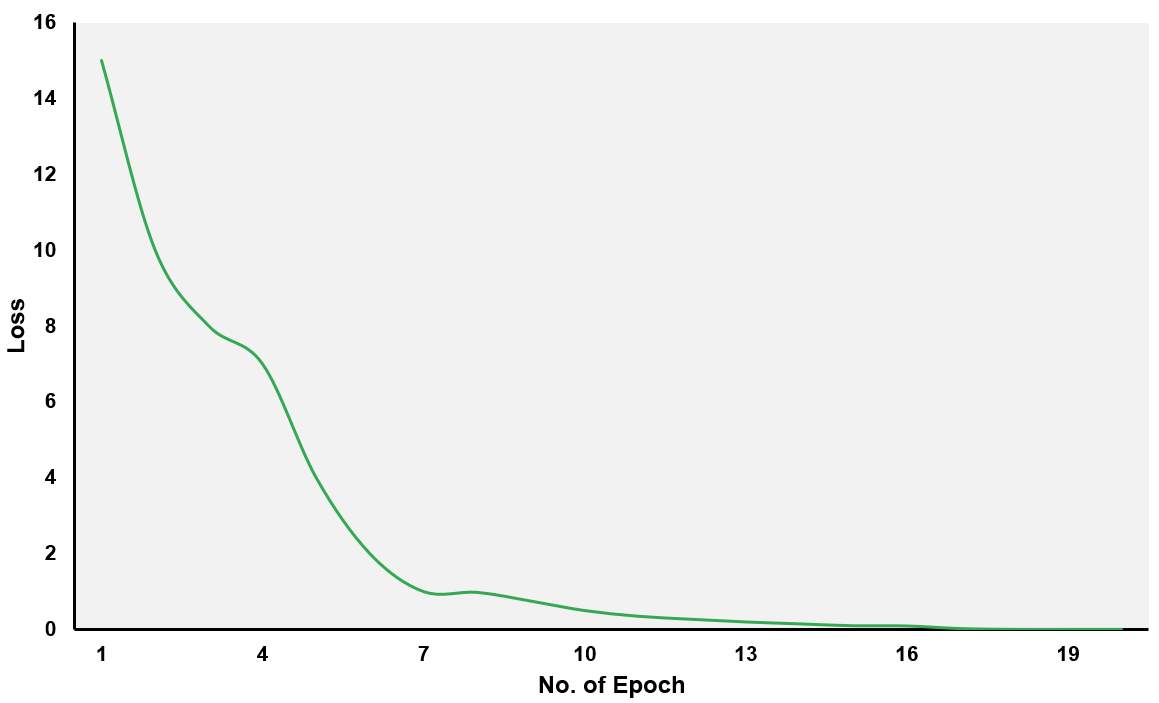}
  \label{loss_dhaka}}
\subfigure[]{
  \includegraphics[width=0.45\linewidth, height=0.27\textwidth]{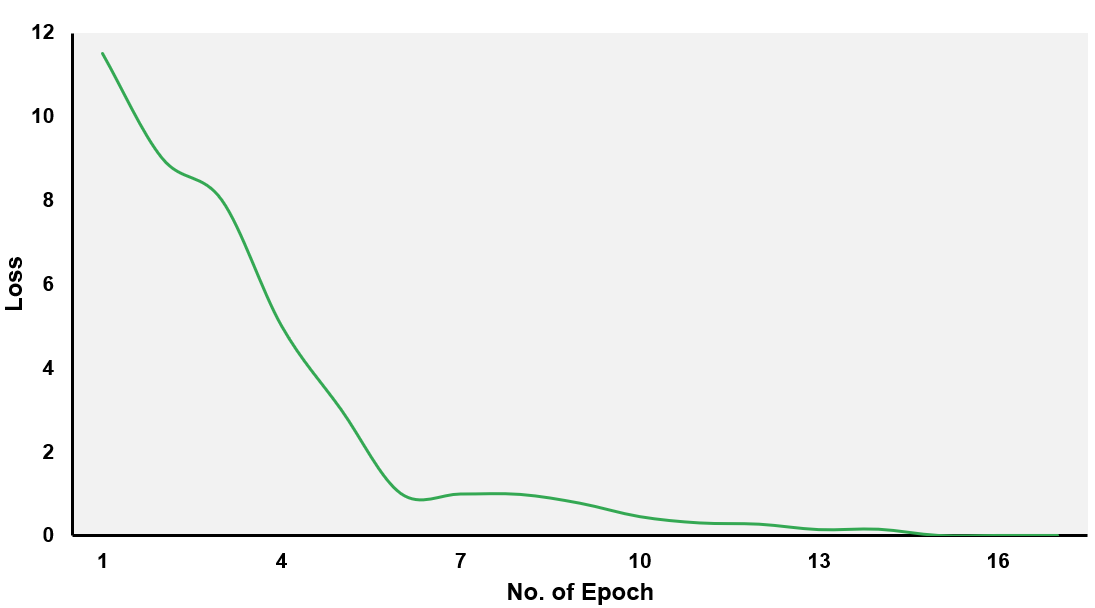}
  \label{loss_raj}}
  
\subfigure[]{
  \includegraphics[width=0.45\linewidth, height=0.27\textwidth]{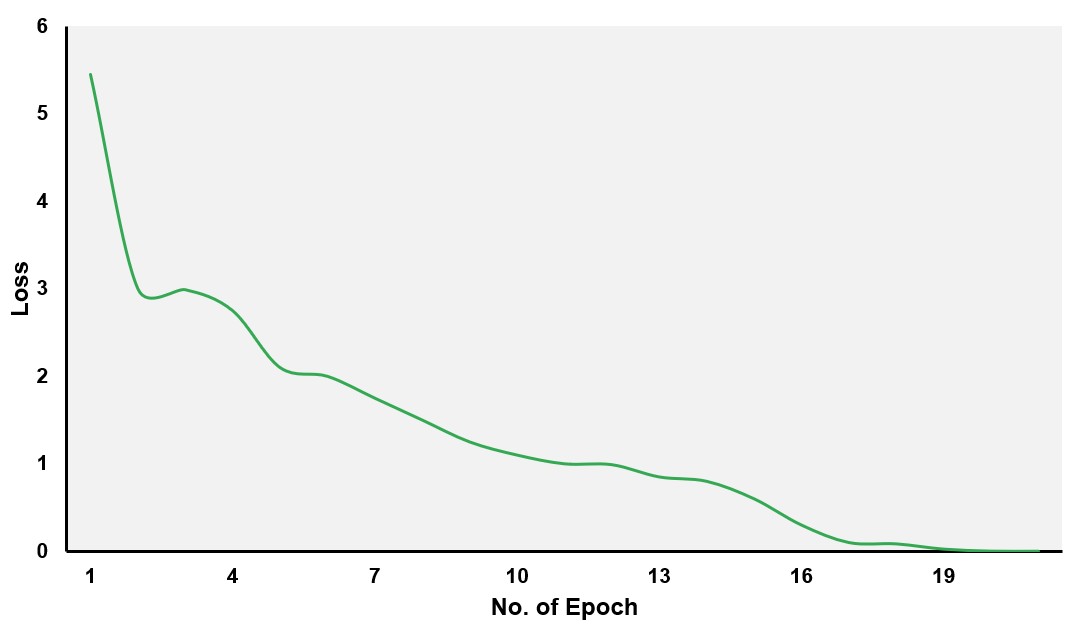}
  \label{loss_raj}}
\subfigure[]{
  \includegraphics[width=0.45\linewidth, height=0.27\textwidth]{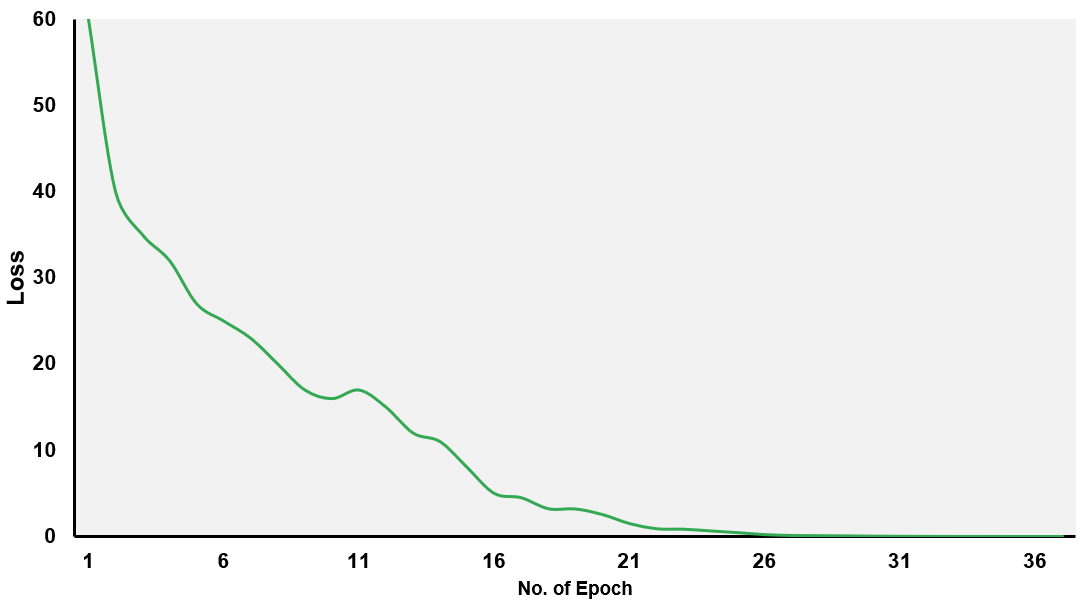}
  \label{loss_khulna}}

\caption{Training loss of the proposed stacked Bi-LSTM models for (a)Dhaka Weather Station, (b)Rajshahi Weather Station, (c)Chittagong Weather Station, and (d)Khulna Weather Station. }
\label{loss}
\end{figure*}

As a forecasting model, we have selected the Bi-LSTM architecture for its performance. After extensive evaluation, we have selected the 3 Bi-LSTM layers with 1 time-distributed layer for the final weather prediction model. The selection process of the preferred model has been vividly discussed in the next subsection \ref{tuning}. We have trained 35 different models for 35 weather stations to cover the entire area of Bangladesh. For demonstration purposes, we have shown four major stations' (Dhaka, Rajshahi, Chittagong, and Khulna)training loss graph in the figure \ref{loss}.
We stopped training the models when the losses were significantly low and remained identical for each epoch.  

\subsubsection{Bi-LSTM Model's Hyperparameter Tuning and Customization }
\label{tuning}
We have selected Bi-LSTM as the best architecture for our forecast model. Now, we are stepping toward the best performance from the model by hyperparameter tuning. We evaluated the model with different hyperparameters, features, and missing value-handling techniques and chose the best one for our deployed model. Table \ref{tab:table16} shows the average $R^{2}$ value for the validation and test set on different features of the different experimental models with the respective missing value handling technique on the data.

\begin{table*}[h!]
\caption{\label{tab:table16}Comparative analysis of different model architectures}
\resizebox{\textwidth}{!}{\begin{tabular}{|c|c|c|c|c|}
\hline
\textbf{Model Architecture} &
  \textbf{Features} &
  \begin{tabular}[c]{@{}l@{}}\bf Average \\ \textbf{R\textsuperscript{2} Value} \\  \bf (Validation)\end{tabular} &
  \begin{tabular}[c]{@{}l@{}}\bf Average\\ \textbf{R\textsuperscript{2} Value} \\  \bf (Test)\end{tabular} &
  \begin{tabular}[c]{@{}l@{}}\textbf{Missing Value }\\ \bf Handling Technique \end{tabular} \\ \hline
\begin{tabular}[c]{@{}l@{}}2 Bi-LSTM layers\\ one-time \\ distributed dense layer\end{tabular} &
  \multirow{5}{*}{\begin{tabular}[c]{@{}l@{}}Humidity, Rainfall, \\ Sunshine, Temperature\end{tabular}} &
  0.959 &
  0.962 &
  \begin{tabular}[c]{@{}l@{}}Forward fill and\\ Backward fill\end{tabular} \\ \cline{1-1} \cline{3-5} 
\begin{tabular}[c]{@{}l@{}}2 Bi-LSTM layers\\ Two-time \\ distributed dense layer\end{tabular} &
   &
  0.9527 &
  0.9567 &
  \begin{tabular}[c]{@{}l@{}}Forward fill and \\ Backward fill\end{tabular} \\ \cline{1-1} \cline{3-5} 
\begin{tabular}[c]{@{}l@{}}\textbf{3 Bi-LSTM layers} \\ \textbf{one-time} \\ \textbf{distributed dense layer}\end{tabular} &
   &
  \textbf{0.9824} &
  \textbf{0.9838} &
  \textbf{Linear interpolation} \\ \cline{1-1} \cline{3-5} 
\begin{tabular}[c]{@{}l@{}}3 Bi-LSTM layers\\ one-time \\ distributed dense layer\end{tabular} &
   &
  0.9524 &
  0.9616 &
  \begin{tabular}[c]{@{}l@{}}Forward fill and \\ Backward fill\end{tabular} \\ \cline{1-1} \cline{3-5} 
\begin{tabular}[c]{@{}l@{}}4 Bi-LSTM layers \\ one-time \\ distributed dense layer\end{tabular} &
   &
  0.9422 &
  0.9546 &
  \begin{tabular}[c]{@{}l@{}}Forward fill and \\ Backward fill\end{tabular} \\ \hline
\begin{tabular}[c]{@{}l@{}}3 Bi-LSTM layers \\ one-time \\ distributed dense layer\end{tabular} &
  \begin{tabular}[c]{@{}l@{}}Humidity, Rainfall, Sunshine, \\ Temperature, Wx, Wy\end{tabular} &
  0.5727 &
  0.6181 &
  Linear interpolation \\ \hline
\begin{tabular}[c]{@{}l@{}}3 Bi-LSTM layers \\ one-time \\ distributed dense layer\end{tabular} &
  \begin{tabular}[c]{@{}l@{}}Humidity, Rainfall, Sunshine, \\ Temperature, Week cos, Year cos\end{tabular} &
  0.4839 &
  0.5350 &
  Linear interpolation \\ \hline
\end{tabular}}

\end{table*}

\begin{itemize}
    \item[--]Firstly, we experimented with the model having 2 Bi-LSTM layers and a time-distributed dense layer. We used four features: Humidity, Rainfall, Sunshine, and Temperature. The average $R^{2}$ value obtained was 0.959 for the validation and 0.962 for the test set. In this case, the missing values were filled using forward and backward fill methods. In the second model, we tuned our model with 2 Bi-LSTM and 2-time distributed dense layers, keeping the features unchanged. The average $R^{2}$ value obtained was 0.9527 for the validation and 0.9567 for the test set.

    \item[--] In the third model, we used 3 Bi-LSTM layers and a time-distributed dense layer, which outperformed all of the other models for the same set of features. In this model, the average $R^{2}$ value was 0.9824 for the validation and 0.9838 for the test set. Here, we filled in the missing value using linear interpolation. As an alternative, we changed the technique of handling the missing values from linear interpolation to the forward-backward filling method. As a result, the average $R^{2}$ value found was 0.9523 for the validation and 0.9616 for the test set.

    \item[--] In the fourth model, we experimented with 4 Bi-LSTM layers, one-time distributed dense layers, and the same features. The average $R^{2}$ value found was 0.9422 for the validation and 0.9546 for the test set. We filled the missing values using forward fill and backward fill. Again, we experimented with 3 Bi-LSTM layers and a time-distributed dense layer. We added new features $W_x$, $W_y$, which are basically wind speed and wind direction converted in vector format. Here, we observed a drastic change in the $R^{2}$ value. The average $R^{2}$ value was 0.5727 for the validation and 0.6181 for the test set.

    \item[--] Moreover, we used the same model but added time (week cosine and year cosine) as a new feature and converted the date to signal so that we could relate the date to other features. The average $R^{2}$ value was 0.4839 for the validation and 0.5350 for the test set. The missing values were handled by linear interpolation.
    
\end{itemize}

\textit{In conclusion, the 3 Bi-LSTM layers and one-time distributed dense layer proved to be the best architecture for the model, and this model achieved an average R\textsuperscript{2} value of 0.9824.} Table \ref{tab:table13} shows the test results of divisional station models for the proposed architecture (3 Bi-LSTM layers and one-time distributed dense layer).

\begin{table}
\caption{\label{tab:table13} Test results of the proposed architecture for the divisional station models.}
\begin{tabular*}{\tblwidth}{|C|C|C|C|C|C|}
\hline
\textbf{City} & \textbf{MAE}    & \textbf{MSE}    & \textbf{MSLE}   & \textbf{R\textsuperscript{2}} & \textbf{SMAPE}  \\ \hline
Barisal & 0.0438 & 0.0018 & 0.0126 & 0.9856               & 0.0279 \\ \hline
Rangpur & 0.0495 & 0.0017 & 0.0129 & 0.9835               & 0.0314 \\ \hline
Khulna & 0.0427 & 0.0017 & 0.0128 & 0.9859               & 0.0296 
\\ \hline
Dhaka & 0.0128 & 0.004 & 0.032 & 0.970                & 0.080 
\\ \hline
Chittagong & 0.041 & 0.002 & 0.013 & 0.985                & 0.027 
\\ \hline
Sylhet      & 0.055 & 0.002 & 0.014 & 0.983                & 0.035 \\ \hline
Rajshahi & 0.073 & 0.002 & 0.013 & 0.984                & 0.048 
\\ \hline
\end{tabular*}
\end{table}

\subsection{Weather Forecast Output and Recommendation}
We have successfully created a weather forecasting model for the betterment of Bangladeshi farmers. The proposed model can generate some decisions for the farmers so that they can be benefited. This section will show the sample output and recommendations for the proposed model.

\subsubsection{Find Nearest Weather Station}
According to the statistics of the Bangladesh government, there are only 35 weather stations to cover 64 districts of Bangladesh \cite{Weatherstation}. So, we want to forecast any location's weather based on the nearest weather station data as described in the previous section \ref{nearest}. For demonstration purposes, we consider a suburb of Bangladesh's capital (Dhaka) named Uttara. Our model will automatically capture the latitude and longitude from the GPS location, which is 23.8759° N and 90.3795° E for Uttara. And Dhaka weather station's latitude and longitude are 23.8111° N and 90.3965° E. Another weather station in the Mymensing district has a latitude and longitude is 24.7471°N, 90.4203°E. If we apply the Haversine formula to find the distance with the help of equation \ref{eu_eqn}, we will get the distance of 7.6 and 96.96 km, respectively. It shows Dhaka weather station is the nearest between the two stations. Thus, we will calculate all the distances from the 35 weather stations of a specific location and select the nearest weather station for our prediction model.   

\subsubsection{Forecast Visualization and Recommendation}
\begin{figure}
  \includegraphics[width=\columnwidth]{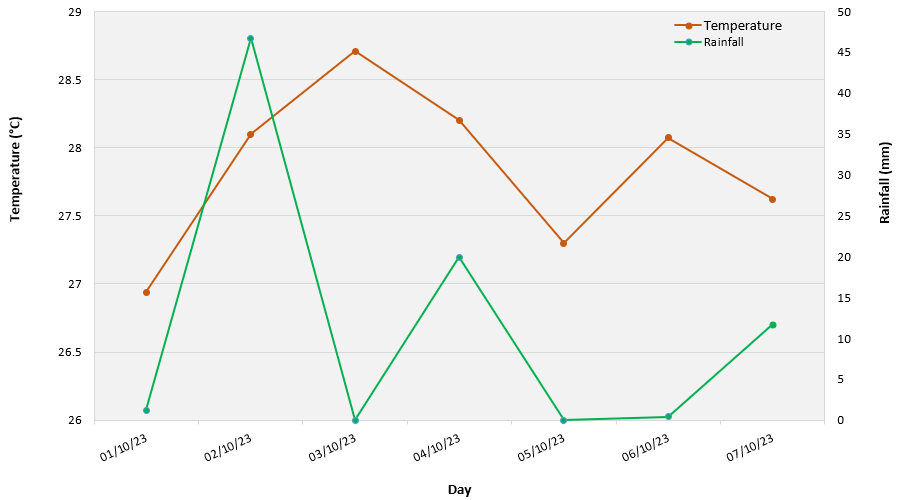}
  \caption{Forecast of Weekly Rainfall and Temperature of Dhaka City.}
  \label{week}
\end{figure}

\begin{figure}
  \includegraphics[width=\columnwidth]{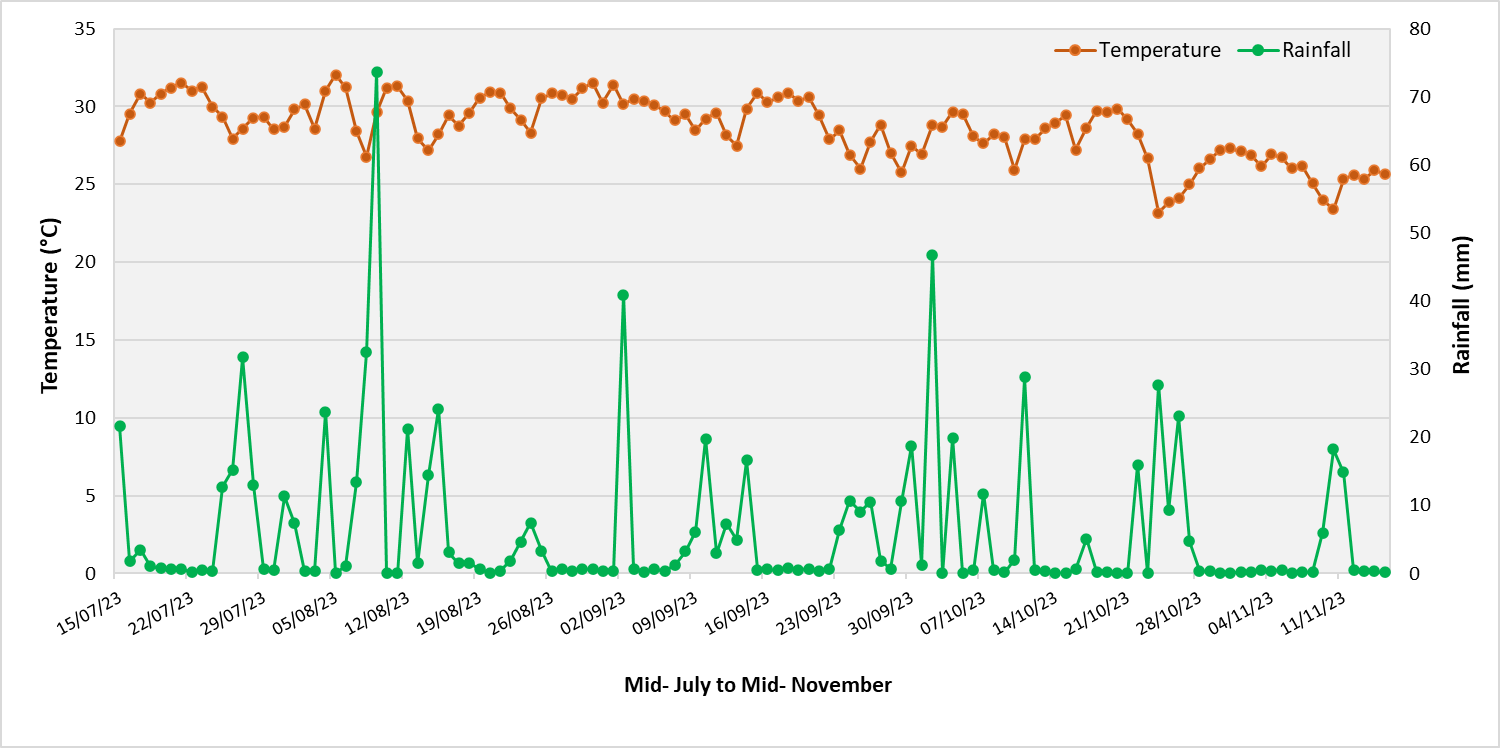}
  \caption{Forecast of Seasonal(Kharif-2) Rainfall and Temperature of Dhaka City. }
  \label{season}
\end{figure}

\begin{figure}
  \includegraphics[width=\columnwidth]{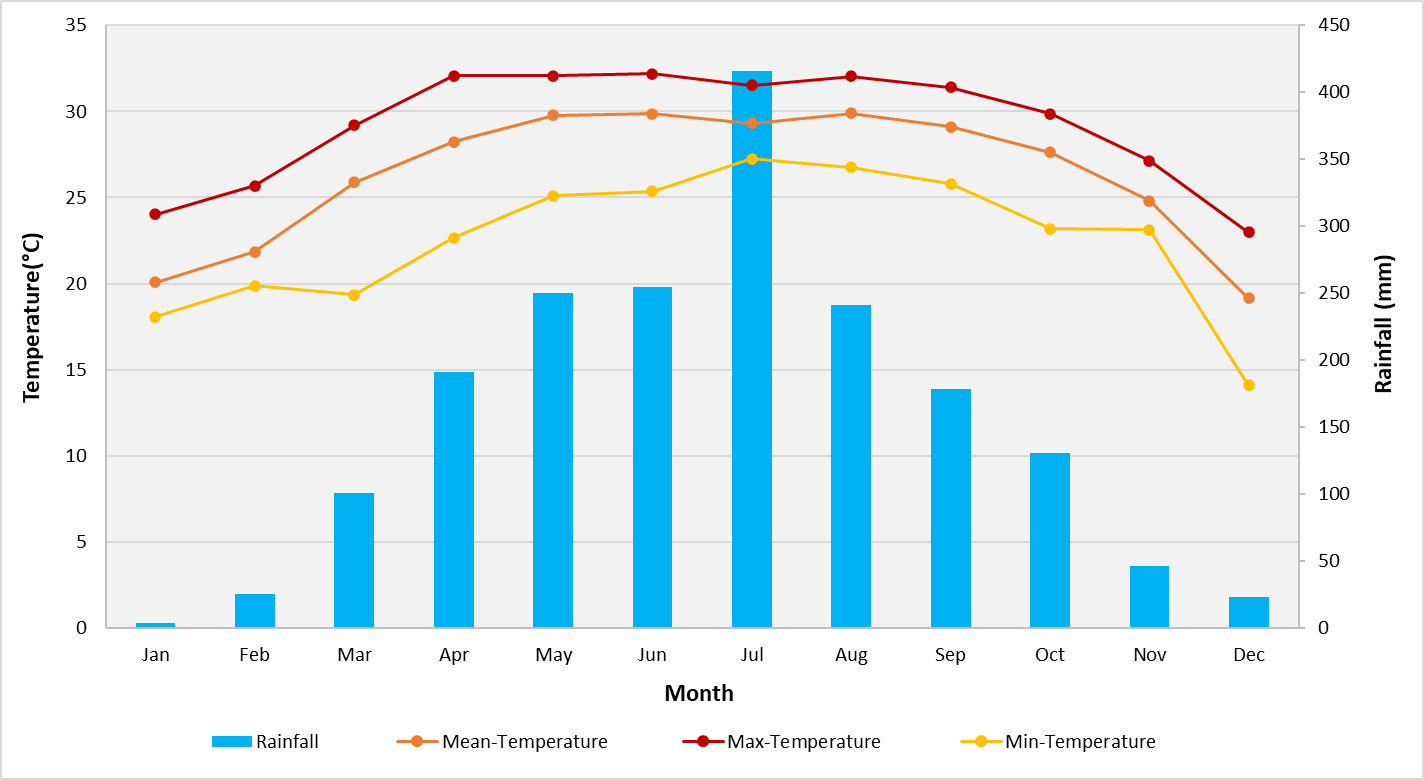}
  \caption{Forecast of Yearly (2023) Rainfall and Temperature of Dhaka City. }
  \label{year}
\end{figure}

From the model, the farmers can effectively get the weather forecast for any future dates within a year. It will help the farmers to make appropriate farming decisions, like sowing seeds, fertilizing, spreading pesticides, harvesting, etc. 
For better understanding, we have included the visual representation of the weather forecast for the Capital City of Bangladesh (Dhaka). Figure \ref{week} shows the weather forecast for the first week of October 2023.   
The line plot in Figure \ref{season} represents the weather forecast for the Kharif-2 season ranging from mid-July to mid-November. With the proposed model, the farmers can also find the visualization for the other two seasons - Rabi and Kharif-1. 
For intuition of the next year's weather, we will show the summary of forecast results as shown in Figure \ref{year}. 
Finally, we will generate automated recommendations for the farmers if the model continuously observes significant changes in the forecast output, especially if the forecast weather becomes significantly lower or higher than the reference threshold value as described in Table \ref{threshold}. Other recommendations are generated as vividly described in the section \ref{recommendation}.

\section{Discussion}
\label{discussion}
Almost half of the population of developing countries (e.g. Bangladesh) relies on agriculture to earn their livelihood. Most often, farmers use conventional methods of farming. Even they aren't aware of climate change, which results in sloppy development in the agricultural sector \cite{development}. This makes our study more significant in the context of agriculture. Previous studies suggest proper knowledge of climate change and weather boosts the production skills of crops. However, farmers are less conscious of meteorological factors \cite{mase2014unrealized}. In many cases, weather forecast output may not provide reliable weather prediction, which discourages farmers.  Therefore, in this paper, we presented a location-based daily, weekly, seasonal, and yearly reliable weather forecast model. Based on this, we also presented an automated recommendation system to assist in the context of agriculture.

Before starting our research work, we had undergone an extensive study to find out whether there was any existing model or not. To the best of our knowledge, this is the first in-depth analysis and recommendation system taking into account real-world data from the perspective of agricultural decision-making (in Bangladesh). However, we faced several challenges. Our first challenge was a huge amount of reliable weather data collection. We collected Bangladeshi weather stations' data with the agreement of Bangladesh's meteorological department. However, the data was unstructured and contained some missing values. We put a lot of effort into the data preprocessing, which is vividly discussed in section \ref{data_collection}. We fed the data to the LSTM and Bi-LSTM models for the forecast. After much evaluation, based on the performance, we selected Bi-LSTM (with 3 Bi-LSTM layers and one time-distributed layer). The selected model achieved a $R^2$ value of 0.9838 on the test data. Finally, we use the model for weather forecasts and agricultural recommendation generation. The model also suggests suitable crops for farmers in low-lying and drought-prone regions. 

Our model is created based on secondary data in an offline learning mode. However, it can be used universally by following the steps of our proposed model, discussed earlier. One may need to train the model with relevant regional weather stations' data. Though we have worked with collected historical data, we plan to collect real-time data, including sensor readings, in the future.
However, there might be cybersecurity issues with misleading information \cite{sarker2024ai}. Therefore, we will extend our analysis from this perspective in our future work.

\section{Conclusion}
\label{conclusion}
In this study, we proposed a parametric weather forecast model trained on the data collected from all 35 weather stations under the Meteorological Department of Bangladesh. The primary goal is to help the farmers in making viable farming decisions using the weather predicted by our weather forecast model. For weather modeling, we experimented with several variants of the Bi-LSTM models. In our extensive experiment, three Bi-LSTM layers along with a one-time distributed dense layer model outperformed other variants by achieving the average R\textsuperscript{2} value of 0.9824 and 0.9838 for validation and testing samples respectively. Lastly, the model displays the weather forecast to the user along with some automated agricultural recommendations. We hope the proposed model will help the farmers to a greater extent. 

% Regarding this, future study should explore sequence modeling framework such as transformers for weather modeling. Moreover, future study should also incorporate weather features such as clouds, visibility, and atmospheric pressure into account for weather modeling.

% In the proposed architecture, we have worked on creating a forecast model for the betterment of Bangladeshi farmers. We have collected the weather data for all 35 weather stations from the meteorological department of Bangladesh. The collected data were unstructured, and there were some missing and outlier values as well. We have used multiple techniques for handling the missing values. In our extensive experiment, three Bi-LSTM layers with a one-time distributed dense layer model outperformed other models, and we picked the model for further forecasting. We have trained the model for each weather station. Lastly, we show the weather forecast to the user of our model, along with some automated agricultural recommendations. We hope the proposed model will help the farmers of Bangladesh to a greater extent. 

\section*{Declaration:} {The authors declare that they have no conflict of interest.}

\bibliographystyle{model1-num-names}
\bibliography{cas-refs}

\end{document}